%% file: acl_latex.tex
\pdfoutput=1

\documentclass[11pt]{article}

\usepackage[final]{acl}

\usepackage{times}
\usepackage{latexsym}

\usepackage[T1]{fontenc}

\usepackage[utf8]{inputenc}

\usepackage{microtype}

\usepackage{inconsolata}

\usepackage{xurl}
\usepackage{booktabs}       
\usepackage{amsfonts}       
\usepackage{nicefrac}       
\usepackage{microtype}      
\usepackage{xcolor}         
\usepackage{graphicx}
\usepackage{wrapfig}
\usepackage{adjustbox}
\usepackage{booktabs}
\usepackage{makecell}
\usepackage[most]{tcolorbox}
\usepackage[framemethod=tikz]{mdframed}
\usepackage{cleveref}
\usepackage{subcaption}
\usepackage{tabularx}
\usepackage{xspace}
\usepackage{multirow}
\usepackage{multicol}
\usepackage[title]{appendix}

\usepackage{tikz}
\usetikzlibrary{tikzmark}

\usepackage{arydshln}

\newcommand{\bluecell}{\cellcolor{blue!10}}

\definecolor{mypink}{RGB}{221, 24, 161}
\definecolor{mygray}{RGB}{174, 182, 191}
\definecolor{mygreen}{RGB}{39, 174, 96}
\definecolor{greenforcaption}{RGB}{180, 215, 167}
\definecolor{pinkforcaption}{RGB}{244, 204, 204}
\definecolor{orangeforcaption}{RGB}{249, 199, 149}

\definecolor{qualcolor}{RGB}{128,64,0}
\gdef\Sepline{%
  \par\noindent\makebox[\linewidth][l]{%
  \hspace*{-\mdflength{innerleftmargin}}%
   \tikz\draw[thick,dashed,gray!60] (0,0) --%
        (\textwidth+\the\mdflength{innerleftmargin}+\the\mdflength{innerrightmargin},0);
  }\par\nobreak}

\newcommand{\oursFramework}{\textsc{DribeR}\xspace}

\newcommand{\drawingEmojiforcaption}{\includegraphics[height=.9em,trim=0 .4em 0 0]{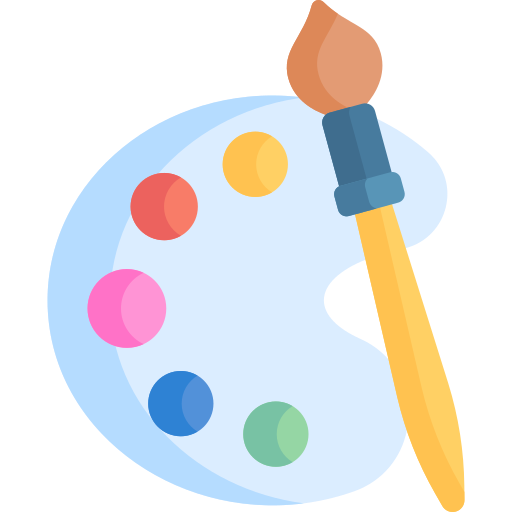}}

\newcommand{\eg}{e.g.,\xspace}
\newcommand{\ie}{i.e.,\xspace}

\newcommand{\photochat}{\textsc{PhotoChat}\xspace}
\newcommand{\photochatplus}{\textsc{PhotoChat++}\xspace}
\newcommand{\augphotochat}{\textit{aug-}\photochat\xspace}

\newcommand{\decision}{\textsc{Decision}$_{\textsc{[Y/N]}}$\xspace}
\newcommand{\intent}{\textsc{Intent}$_{\textsc{[Choice]}}$\xspace}
\newcommand{\evidence}{\textsc{Sentence}$_{\textsc{[Dist.]}}$\xspace}

\title{Large Language Models can \textit{Share} Images, Too!}

\author{Young-Jun Lee\textsuperscript{\rm 1} \hspace{0.3cm}
        Dokyong Lee \textsuperscript{\rm 2} \hspace{0.3cm}
        Joo Won Sung \textsuperscript{\rm 2} \hspace{0.3cm}
        Jonghwan Hyeon \textsuperscript{\rm 1} \hspace{0.3cm}
        Ho-Jin Choi \textsuperscript{\rm 1}\\
    \textsuperscript{\rm 1} School of Computing, KAIST \hspace{0.3cm}
    \textsuperscript{\rm 2} KT Corporation \hspace{0.3cm} \\
    \texttt{\{yj2961, jonghwanhyeon, hojinc\}@kaist.ac.kr} 
    \hspace {0.3cm} \texttt{\{dokyong.lee, jwsung\}@kt.com}
}

\begin{document}
\maketitle

\input{sections/00_abstract}

\input{sections/01_introduction}

\input{sections/02_photochat++}

\input{sections/03_method}
\input{sections/04_experimental_setup}
\input{sections/05_experiment_result}
\input{sections/06_related_work}
\input{sections/07_conclusion}

\section*{Acknowledgement}

This work was supported by a grant of the KAIST-KT joint research project through AI Tech Lab, Institute of convergence Technology, funded by KT [Project No. G01230605, Development of Task-oriented Persona-based Dialogue Generation Combining Multi-modal Interaction and Knowledge Modeling].

\bibliography{custom}

\clearpage

\appendix

\section{Details of \photochatplus} \label{supp:intent_expl}

We describe each intent category as follows:

\paragraph{(1) Information Dissemination.} This intent involves sharing images that convey important information, such as current news, economic updates, educational content, or plot summaries. These images are intended to inform or educate the recipient.
\paragraph{(2) Social Bonding.} This category encompasses sharing personal photographs, memories of recent encounters, recollections of past events (\ie memory recall), or images that connect to one's own experiences. Such image sharing is primarily used to strengthen social ties.   
\paragraph{(3) Humor and Entertainment.} This intent involves sharing images that aim to amuse or entertain, like humorous pictures or memes. The primary focus is on sharing light-hearted content to bring joy or laughter.
\paragraph{(4) Visual Clarification.} Here, images serve as supplementary material to clarify complex concepts or situations, especially when textual conversation alone is insufficient. Images can significantly enhance understanding in such contexts.
\paragraph{(5) Topic Transition.} In this category, images are utilized to change the topic of conversation or modify the mood subtly. This can be a strategic approach to redirecting a discussion or lightening the atmosphere.
\paragraph{(6) Expression of Emotion or Opinion.} This intent includes the sharing of images to express emotions, opinions, or reactions that are challenging to communicate through text alone. It involves the use of emotive photography, art, or reaction images that effectively convey feelings or viewpoints.

\section{Comparison to Previous System}

\begin{figure*}[!t]
    \centering
    \includegraphics[width=0.9\textwidth]{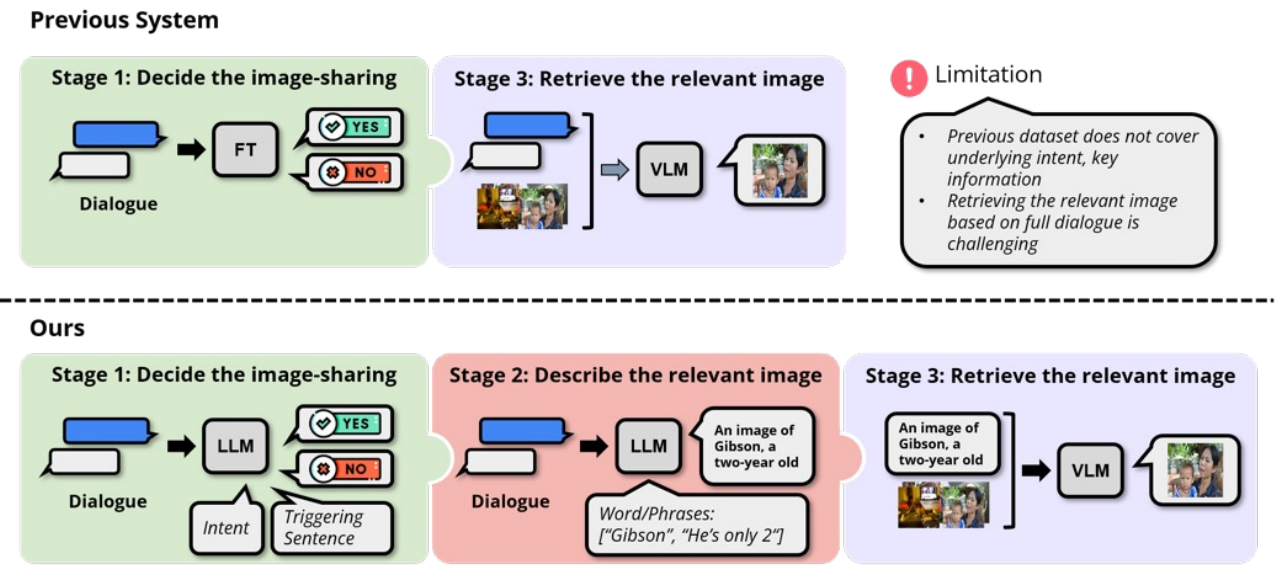}
    \caption{We compare \oursFramework with the previous systems. FT denotes the fine-tuned model as shown in Table~\ref{tab:decision_main_result}.}
    \label{supp_fig:comp_framework}
    \vspace{-1em}
\end{figure*}

Figure~\ref{supp_fig:comp_framework} illustrates the overview of previous systems~\cite{zang2021photochat,feng2022mmdialog} and \oursFramework.

\section{Additional Experiments}

\begin{figure}[!t]
    \centering
    \includegraphics[width=0.9\columnwidth]{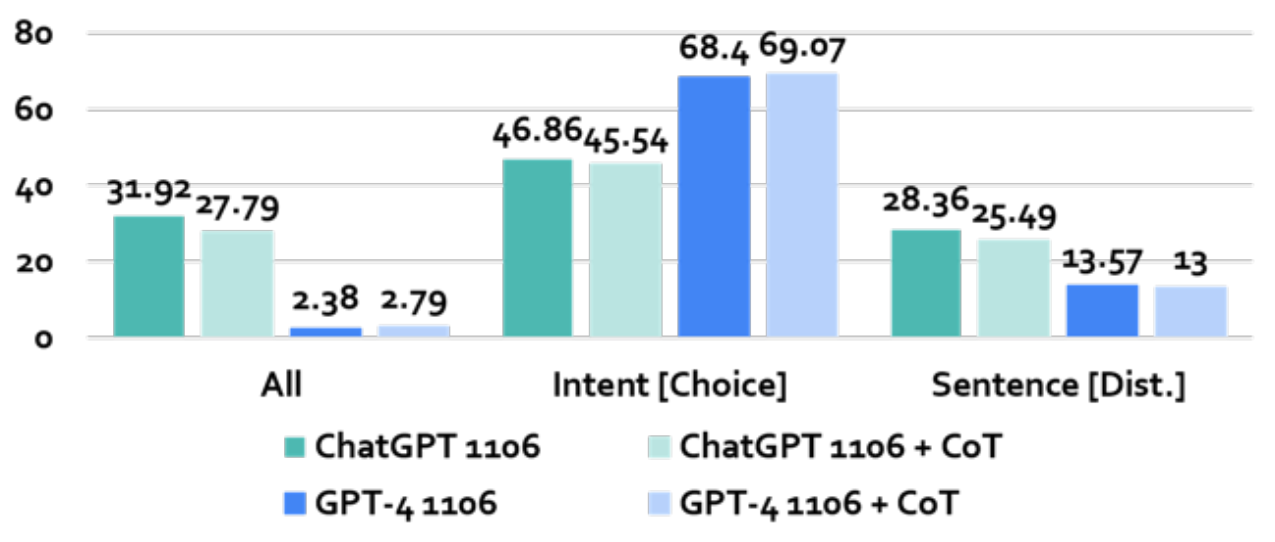}
    \caption{Results of \intent, \evidence, and \textsc{All} when CoT is applied.}
    \label{supp_fig:cot_intent}
    
\end{figure}

Figure~\ref{supp_fig:cot_intent} shows the zero-shot results of \intent, \evidence, and \textsc{All} when we apply zero-shot chain-of-thought (CoT) reasoning. For CoT, we follow \cite{kojima2022large} and use the prompt ``\textit{Let's think step by step}''. We observe that CoT does not improve the performance of \intent, \evidence, and \textsc{All}. These results suggest that CoT selectively improves the image-sharing capability of LLMs.

\input{sections/supplementary/00_human-evaluation}
\input{sections/supplementary/01_intrinsic}
\input{sections/supplementary/02_implementation_details}
\input{sections/supplementary/03_rationale_dist}
\input{sections/supplementary/04_template}

\input{sections/supplementary/05_case_studies}

\end{document}

%% file: sections/00_abstract.tex
\begin{abstract}

This paper explores the \textit{image-sharing} capability of Large Language Models (LLMs), such as GPT-4 and LLaMA 2, in a zero-shot setting. To facilitate a comprehensive evaluation of LLMs, we introduce the \photochatplus dataset, which includes enriched annotations (\ie \textit{intent}, \textit{triggering sentence}, \textit{image description}, and \textit{salient information}). Furthermore, we present the gradient-free and extensible \textbf{D}ecide, Desc\textbf{ribe}, and \textbf{R}etrieve (\oursFramework) framework. 
With extensive experiments, we unlock the \textit{image-sharing} capability of \oursFramework equipped with LLMs in zero-shot prompting, with ChatGPT achieving the best performance.
Our findings also reveal the emergent \textit{image-sharing} ability in LLMs under zero-shot conditions, validating the effectiveness of \oursFramework. We use this framework to demonstrate its practicality and effectiveness in two real-world scenarios: (1) human-bot interaction and (2) dataset augmentation. To the best of our knowledge, this is the first study to assess the \textit{image-sharing} ability of various LLMs in a zero-shot setting.
We make our source code and dataset publicly available~\footnote{\url{https://github.com/passing2961/DribeR}}.

\end{abstract}

%% file: sections/01_introduction.tex
\section{Introduction} \label{sec:intro}

\begin{figure}[!t]
    \centering
    \includegraphics[width=0.7\columnwidth]{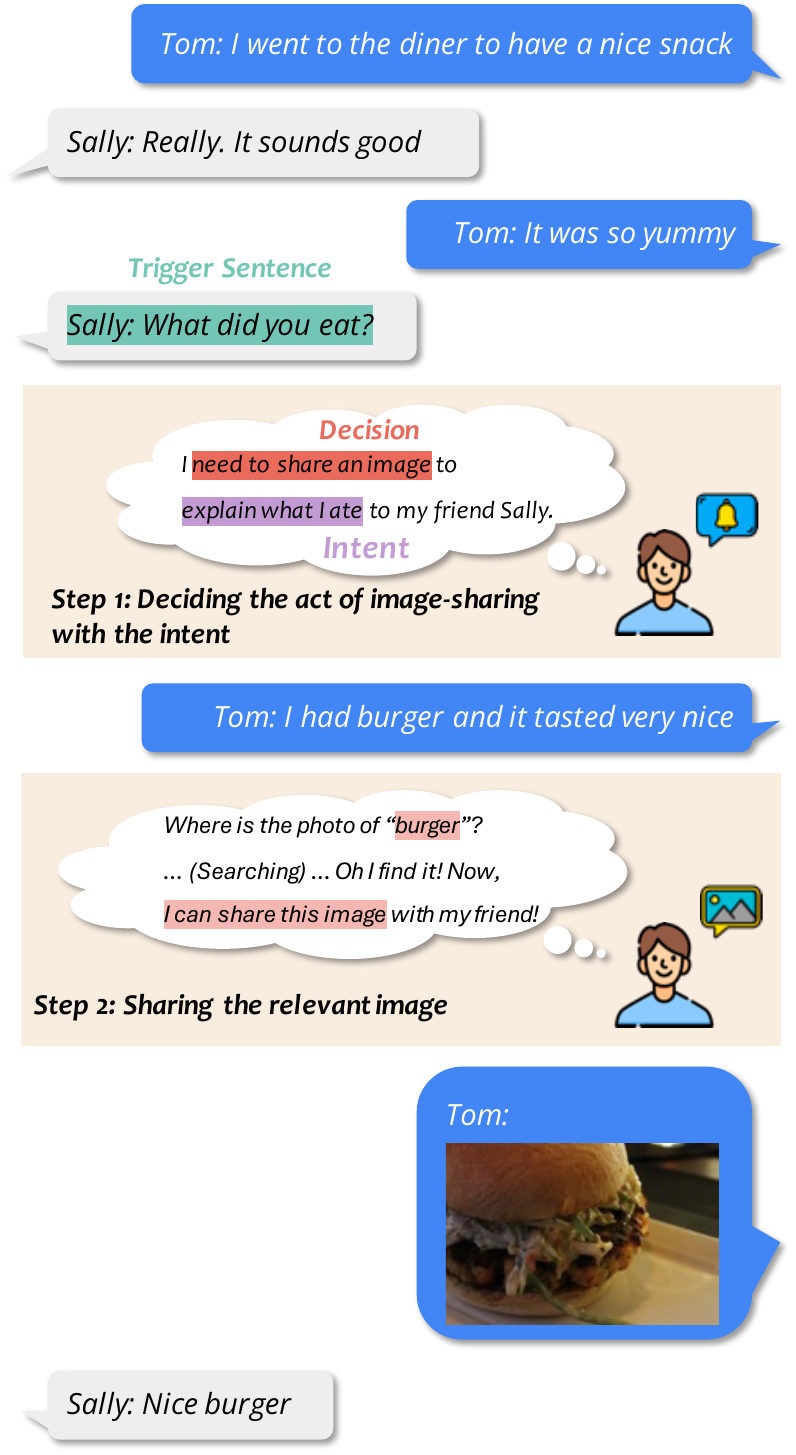}
    \caption{An illustration of human's internal two-stage system of the image-sharing behavior.}
    \label{fig:teaser}
    \vspace{-1em}
\end{figure}

People often share a variety of images during interactions via instant messaging tools.
In practice theory, this is referred to as \textit{photo-sharing} behavior~\citep{lobinger2016photographs}, which is interpreted as a communicative practice.~\footnote{From now on, we refer to this as \textit{image-sharing} behavior, given that ``image'' is a broader concept than ``photo,'' thereby providing more flexibility to language models.}
This behavior operates internally through a two-stage system~\cite{zang2021photochat}: (1) \textit{when to share} and (2) \textit{what to share}. For example, as shown in Figure~\ref{fig:teaser}, we first discern the appropriate moment (\ie decision) for sharing an image with certain intent based on one sentence that invokes the image-sharing behavior.
Then, we share a relevant image at that moment, either by searching the internet or using photos taken on our mobile devices.
This work primarily focuses on unlocking the image-sharing behavior capabilities of Large Language Models (LLMs) in a zero-shot manner.

However, recent studies related to multi-modal dialogue exhibit two significant limitations; \textbf{(1) Over-Simplification.} As shown in Figure~\ref{fig:teaser}, we humans decide to share images with an underlying intent, such as \textit{visual clarification}, due to a certain triggering sentence (\eg ``\textit{What did you eat?}'') in the previous dialogue context. Nevertheless, existing studies~\cite{zang2021photochat,feng2022mmdialog} have reduced this complex behavior to a binary format (\ie ``yes'' or ``no''), which oversimplifies the complexity of image-sharing behavior. This simplification limits a comprehensive understanding of image-sharing behavior. 
(2) \textbf{Limited Understanding of Dialogue.} 
Since the key evidence is scattered across the entire dialogue context~\cite{chae2023dialogue}, understanding the linguistic cues underlying dialogue context is critical~\cite{wang2023cue} for retrieving images relevant to the dialogue.
Nevertheless, existing systems~\cite{lee2021constructing,zang2021photochat,feng2022mmdialog} have primarily leveraged a dialogue-image matching score based on the cosine similarity between the two modalities (\ie dialogue and image) to retrieve the relevant image at the image-sharing moment. This approach leads to a lack of capacity to comprehend the dialogue context due to the limitation of dual encoder structures (\eg CLIP) in fully grasping the dialogue context~\cite{yin2024dialclip}. 

This work explores whether LLMs contain the \textit{image-sharing} capability, primarily focusing on a zero-shot performance. 
To this end, we introduce \textbf{D}ecide, Desc\textbf{ribe}, and \textbf{R}etrieve (\oursFramework), a gradient-free, extensible, and generalizable framework to unlock this \textit{image-sharing} capability of LLMs through in-context zero-shot learning. 
Broadly, \oursFramework consists of three stages: (1) deciding the image-sharing behavior with the intent, (2) describing the image description relevant to the previous dialogue, and (3) retrieving the relevant image to the image description. The overall pipeline is illustrated in Figure~\ref{fig:framework}.
In addition, we introduce \photochatplus, an extended version of \photochat. \photochat is a multi-modal dialogue dataset constructed via crowdsourcing for human-human interaction. \photochatplus contains six intent labels, a triggering sentence, and salient information (\eg ``words'' or ``phrases'') to invoke the image-sharing behavior.
In our experiments, we successfully unlock the \textit{image-sharing} capability of LLMs in a zero-shot setting with the aid of \oursFramework, with ChatGPT achieving state-of-the-art performance. Using the \photochatplus dataset, we demonstrate that image-sharing is a challenging task for both humans and LLMs even if we apply the few-shot setting and Chain-of-Thought reasoning. Our extensive experiments further confirm that our framework is effective and versatile in real-world applications, specifically in (1) human-bot interaction dialogue and (2) dataset augmentation.

In summary, our main contributions are as follows:
1) We introduce the Decide, Describe, Retrieve (\oursFramework) framework, designed to evaluate the \textit{image-sharing} ability of LLMs in a zero-shot setting.
2) For a comprehensive assessment of LLMs' image-sharing capabilities, we present the \photochat dataset, enriched with additional information such as intent, triggering sentences, image descriptions, and salient information.
3) Compared to the existing method, Experimental results show that \oursFramework with LLMs achieves competitive zero-shot performance, even without additional training.
4) To the best of our knowledge, this is the first study to test the \textit{image-sharing} capability of LLMs through zero-shot prompting.

%% file: sections/02_photochat++.tex
\section{Overview of \photochatplus} \label{sec:photochat++}

The \photochatplus provides additional information (\ie intent, triggering sentence, image description, and salient information) related to the image-sharing behavior by annotating the original \photochat test set. The purpose of this dataset is to thoroughly assess the image-sharing capability of LLMs based on the internal operation system of humans. We describe the details of the human annotation procedure in Appendix~\ref{supp:annotation_procedure}.

\subsection{Preliminary: \photochat} 

The \photochat~\citep{zang2021photochat} is a human-human multi-modal dialogue dataset constructed through crowdsourcing. 
This dataset contains 10k multi-modal dialogues, where each dialogue $\mathcal{D}=\{(u_1, s_1), ..., (u_{t-1}, s_{t-1}), (i_t, s_t), (u_{t+1}, s_{t+1}), \allowbreak ..., (u_N, s_N)\}$ in the dataset contains only one image $i_t$ to be shared at turn $t$. 
The $N$ and $s_j \in \{0, 1\}$ denote the number of dialogue turns and speaker information, respectively. 
In addition, they define two tasks by decomposing the image-sharing behavior --- a photo-sharing intent prediction task and an image retrieval task. The formulations are described as follows.

\paragraph{Task 1: Photo-Sharing Decision Prediction.} Given the dialogue history $(u_j)_1^{t-1}$ and the corresponding speaker information $(s_j)_1^{t-1}$, this task aims to predict whether it is appropriate to share the image at turn $t$ in the binary classification formulation, where the label $y \in \{0, 1\}$.~\footnote{Originally, this task is called as ``photo-sharing intent prediction''. However, we consider that the ``decision'' term is more suitable than the term ``intent'' in this task because this task just predicts ``\textit{yes}'' or ``\textit{no}''.}

\paragraph{Task 2: Image Retrieval.} Given the dialogue history $(u_j)_1^{t-1}$ and the corresponding speaker information $(s_j)_1^{t-1}$, this task aims to retrieve most appropriate image at turn $t$ from the image candidate set.

\input{tables/main_tables/intent_taxonomy}

\subsection{Intent Category}
As shown in Table~\ref{tab:intent_taxonomy}, we design six intent labels for image-sharing behavior: Information Dissemination, Social Bonding, Humor and Entertainment, Visual Clarification, Topic Transition, and Expression of Emotions or Opinions. The detailed explanation is described in Appendix~\ref{supp:intent_expl}.

\subsection{Collecting Annotations from Humans}

We collect additional information through the human annotation process based on the following considerations.
\textbf{(1) Intent.} Recognizing that the predefined intents for image-sharing behavior are not always mutually exclusive and can intersect based on the context and content of the image, we instruct annotators to select all intents that are applicable. 
\textbf{(2) Triggering Sentence.} Annotators are asked to identify and highlight the most significant sentence (\ie only one sentence) from the preceding dialogue that contributed to the decision to share an image. 
\textbf{(3) Image Description.} We request annotators to write only one image description, beginning with prefix phrases such as ``An image of'' or ``A photo of''. This format aims to standardize the descriptions for ease of analysis.
\textbf{(4) Salient Information.} We ask annotators to highlight all words or phrases they focus on in generating the image description. This step is crucial for understanding the key elements that influence the description process in the human internal mind.

\begin{figure}[t]
    \centering
    \includegraphics[width=\linewidth]{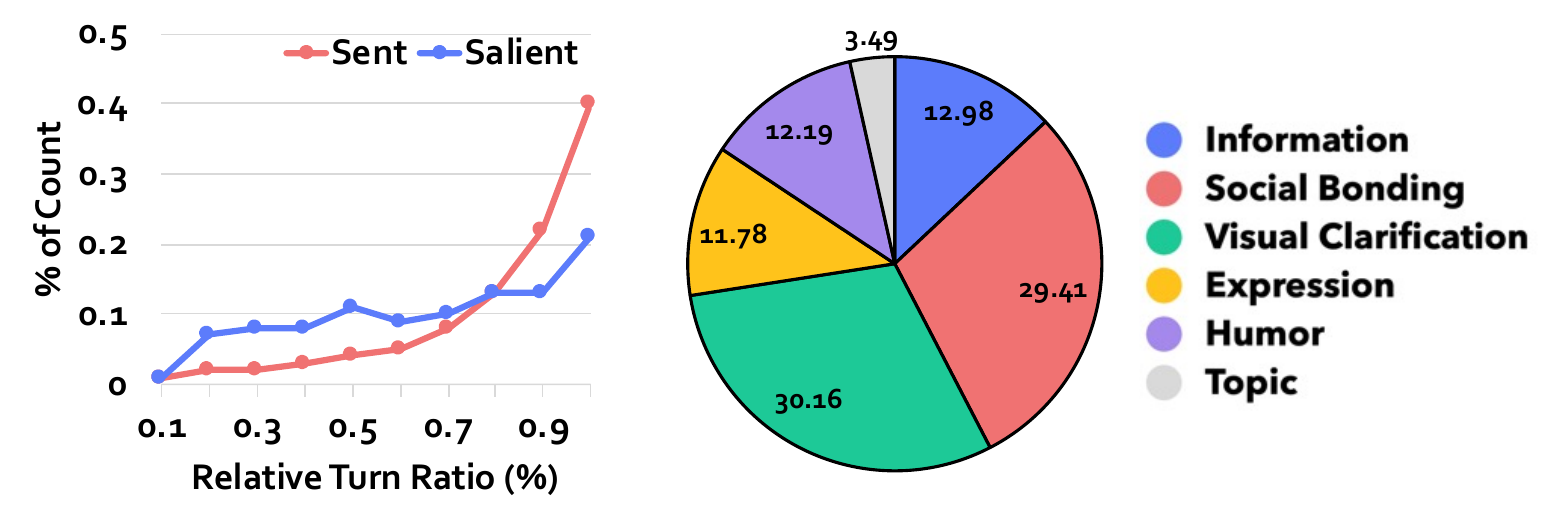}
    \caption{Analysis of \photochatplus. (Left) the distribution of triggering sentence and salient information. (Right) the intent distribution.}
    \label{fig:photochat_plus_stat}
    \vspace{-1em}
\end{figure}

\begin{figure*}[!t]
    \centering
    \includegraphics[width=\textwidth]{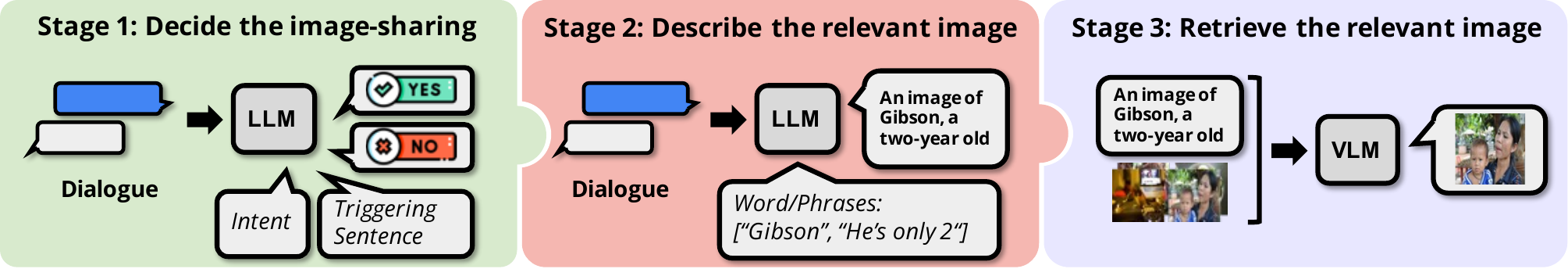}
    \caption{An illustration of our proposed framework: \textbf{D}ecide, Desc\textbf{ribe}, and \textbf{R}etrieve (\oursFramework)}
    \label{fig:framework}
    \vspace{-1em}
\end{figure*}

\subsection{Analysis of \photochatplus}

We analyze \photochatplus, focusing on intent distribution and the distribution of triggering sentences and salient information.

\paragraph{Intent Distribution.}
Visual clarification is the most prevalent, indicating that images are often used in social conversations to aid understanding. Social bonding was also common, likely because the \photochat is designed for social bonding. The least common is Topic transition, which is logical since topic transitions are more likely in longer conversations. However, the \photochat typically has only 12 utterances per conversation, shorter than many long-term dialogue datasets. Topic transition is interesting and related to proactiveness, suggesting that the future creation of long-term multi-modal dialogue datasets could be beneficial.

\paragraph{Distribution of Key Information.}

We analyze which utterances in the dialogue just before image sharing trigger this behavior and where in the previous dialogue people focus when creating image descriptions.
It's evident that image-sharing behavior is often triggered by utterances immediately preceding the image-share. Notably, the words or phrases that people cognitively focus on when creating image descriptions are not only distributed in the immediate preceding utterance but also throughout the early and middle parts of the conversation. This shows the importance of a model's ability to understand the entire dialogue when performing dialogue-to-image retrieval.

%% file: tables/main_tables/intent_taxonomy.tex
\begin{table}[t]
    \centering
    \begin{adjustbox}{width=\columnwidth}
    
    \begin{tabular}{@{}ll@{}}
    \toprule
    Intent                           & Purpose                                     \\ \midrule
    Information Dissemination        & To convey important information             \\
    Social Bonding                   & To strengthen social relationship           \\
    Humor and Entertainment          & To amuse or entertain                       \\
    Visual Clarification             & To clarify complex situations               \\
    Topic Transition                 & To change the topic of dialogue             \\
    Expression of Emotion or Opinion & To express emotions, opinions, or reactions \\ \bottomrule
    \end{tabular}
    \end{adjustbox}
    \caption{Intent Category for Image-Sharing Behavior.}
    \label{tab:intent_taxonomy}
\end{table}

%% file: sections/03_method.tex
\section{\oursFramework} \label{sec:method}

\subsection{Input Prompt Template}

To elicit the \textit{image-sharing} ability of LLM in a zero-shot setting, we manually construct a QA-style prompt template for our framework. The prompt template consists of four main parts: \texttt{[instruction]}, \texttt{[dialogue]}, \texttt{[restrictions]}, and \texttt{[question]}. For each stage, we use different sentences for \texttt{[instruction]}, \texttt{[restrictions]}, and \texttt{[question]}. To explicitly control the model's output, we add \texttt{[restrictions]} with \texttt{Restrictions:}.
In addition, to make the \texttt{[dialogue]} more natural, we replace $s_j$ with Top-1K common names of US SSN applicants from 1990 to 2021~\footnote{\href{https://catalog.data.gov/dataset/baby-names-from-social-security-card-applications-national-data}{\nolinkurl{catalog.data.gov/dataset/baby-names-from-social-security-card-applications-national-data}}}, followed by a previous work~\cite{kim2022soda}. We present the prompt template in Appendix.

\subsection{Stage 1: Deciding Image-Sharing}

Given the dialogue history $(u_j)_1^{t-1}$, the corresponding speaker information $(s_j)_1^{t-1}$, prompt $p_1$ and model $\mathcal{M}$, this stage generates an output $\hat{y_o}$ about whether it is appropriate to share the image: $\hat{y_o} = \mathcal{M}(p_1)$.
Previous studies~\cite{zang2021photochat,li2023pace} ask the model $\mathcal{M}$ to predict the decision, through a simple binary response (\textit{yes} or \textit{no}), whether sharing an image is proper or not.
In this work, we consider additional information, \textit{intent} and \textit{triggering sentence}, to investigate the image-sharing capability of the model $\mathcal{M}$ thoroughly. 
Specifically, the model $\mathcal{M}$ predicts all possible intents among the multiple-choice options and generates one sentence that invokes the image-sharing behavior for the same dialogue.
When the predicted decision is ``\textit{yes}'', then we generate the relevant image description for the same dialogue in stage 2.

\subsection{Stage 2: Describing Relevant Image}

Next, stage 2 uses the same model $\mathcal{M}$ to generate the image description $\hat{y}_{\text{desc}}$ conditioned on the previous dialogue history $(u_j)_1^{t-1}$: $\hat{y}_{\text{desc}} = \mathcal{M}(p_2)$.
Then, instead of using the dialogue $(u_j)_1^{t-1}$ to retrieve the image, as in previous works, we use $\hat{y}_{\text{desc}}$ as a query to retrieve the relevant image. Therefore, the result of retrieved image is bound to depend on the quality of the corresponding $\hat{y}_{\text{desc}}$.

\subsection{Stage 3: Retrieving Relevant Image}

This stage aims to retrieve the relevant image based on the generated image description ($\hat{y}_{\text{desc}}$) from stage 2 by leveraging the vision-and-language pre-trained model (VLM). In this work, we use VLM as CLIP~\cite{radford2021learning}, which is a well-generalized and widely used model in various multi-modal tasks. However, our framework could work with any VLMs, such as BLIP~\cite{li2022blip} or ALIGN~\cite{jia2021scaling}.

\section{Evaluating Image-Sharing Capability}

To verify the performance for each stage, we measure the performance on various evaluation metrics. Each metrics are described as follows.

\subsection{Evaluation for Stage 1}
To understand how well the model $\mathcal{M}$ predicts the decision, intent, and triggering sentence, we report three types of metrics:

\paragraph{(1) \decision.} The model should predict the decision with ``\textit{yes}'' or ``\textit{no}''. We measure the macro F1 score between the ground-truth decision label $y_{\text{D}}$ and the predicted one $\hat{y}_{\text{D}}$.
\vspace{-0.5em}
\paragraph{(2) \intent.} The model should choose all possible intents among the multiple-choice options. Given the ground-truth intents $y_{\text{I}}$ and the predicted intents $\hat{y}_{\text{I}}$, we measure the F1 score between $y_{\text{I}}$ and $\hat{y}_{\text{I}}$. 
\vspace{-0.5em}
\paragraph{(3) \evidence.} The model should generate one most contributed sentence to the image-sharing behavior in a free-form response. We measure the distance between the ground-truth sentence $y_{\text{S}}$ (in \photochatplus) and the predicted response $\hat{y}_{\text{S}}$ by using the token F1 score. Specifically, we get the averaged token F1 score between $\hat{y}_{\text{S}}$ and the ground-truth sentences in \photochatplus.

\subsection{Evaluation for Stage 2}
Based on the theoretical view~\cite{jaimes1999conceptual,santurkar2022caption}, we evaluate the image description  in terms of descriptiveness and completeness. In addition, we evaluate whether the model has a similar cognitive process to a human with respect to salient information. We report three types of metrics:

\vspace{-0.5em}

\paragraph{(1) \textsc{Descriptiveness}}~\cite{santurkar2022caption}. This measures the inter-modal consistency: how much the image description ($\hat{y}_{\text{desc}}$) can replace the ground-truth image provided by \photochat. However, measuring the descriptiveness is infeasible. Thus, previous work~\cite{santurkar2022caption} approximates it with the help of an image-text matching model (\eg BLIP~\cite{li2023blip}). Here, we use CLIPScore~\cite{hessel2021clipscore}. 
\vspace{-0.8em}

\paragraph{(2) \textsc{Completeness}}~\cite{santurkar2022caption}. This involves how much the generated image description ($\hat{y}_{\text{desc}}$) represents the object in the image. We measure the intersection ratio between the object $\hat{y}_{\text{obj}}$ in $\hat{y}_{\text{desc}}$ and the object $y_{\text{obj}}$ in the ground-truth image is $\frac{|y_{\text{obj}} \cap \hat{y}_{\text{obj}}|}{|y_{\text{obj}}|}$. While $y_{\text{obj}}$ is provided by the \photochat, we extract the object $\hat{y}_{\text{obj}}$ in $y_{\text{desc}}$ by prompting ChatGPT to extract the object in the given object categories from \photochat. The prompt we used is presented in Appendix.
\vspace{-0.5em}
\paragraph{(3) \textsc{Consistency.}} This measures the intra-modal consistency between the generated image description ($\hat{y}_{\text{desc}}$) and the human-written image description in \photochatplus. We measure the averaged sentence similarity using Sentence-BERT~\cite{reimers2019sentence}~\footnote{We use sentence-transformers/all-roberta-large-v1 model.}.
\vspace{-0.5em}
\paragraph{(4) \textsc{Salient Information}.} To understand whether LLMs show similar cognitive processes when they generate image descriptions, we ask the model $\mathcal{M}$ to generate the salient words or phrases that are used to generate the image description. We measure the average token F1 score between the model prediction and ground-truth salient information in \photochatplus.

\vspace{-0.5em}

\subsection{Evaluation for Stage 3}

We evaluate how the generated image description retrieves relevant images using the CLIP model. We use a standard metric Recall@$\{1, 5, 10\}$ and mean reciprocal rank (MRR).

\vspace{-0.5em}

%% file: sections/04_experimental_setup.tex
\section{Experiments} \label{sec:expr_setup}

\subsection{Large Language Models in \oursFramework}
The primary objective is to assess the \textit{image-sharing} capability of \oursFramework equipped with LLM in terms of zero-shot performance, which necessitates complex reasoning. 
To achieve this, it is inevitable to leverage instruction-tuned large language models.
For proprietary LLMs, we evaluate 2 models in total: 1) ChatGPT~\citep{chatgpt}, and 2) GPT-4~\citep{openai2023gpt}.~\footnote{We conduct experiments with all language models by calling the OpenAI API between November-2023 and December-2023.}.
For open-sourced LLMs, we evaluate 3 models in total: 1) \textsc{Vicuna} 13B~\citep{chiang2023vicuna}, 2) \textsc{Falcon Instruct} (40B; ~\citep{wang2023far}), and 3) \textsc{LLaMA2 Chat 70B}~\citep{touvron2023llama}.
We present the hyperparameter settings for each stage in Appendix~\ref{supp_sec:implementation}.

%% file: sections/05_experiment_result.tex
\subsection{Results} \label{sec:result}

\input{tables/main_tables/decision_main_result}

\paragraph{\oursFramework unlocks the image-sharing capability.}
Table~\ref{tab:decision_main_result} shows the zero-shot results of various models of \decision on \photochat. 
Compared to the fine-tuned models, we mainly show that most LLMs can share images by understanding the given dialogue context without additional training on \photochat. This is notable when compared to the performance of models that have been fine-tuned. Such a finding suggests that \oursFramework can be effectively utilized in social dialogues requiring an understanding of, and an ability to imagine, interactions between multiple individuals, which is benefited from the power of instruction-tuned LLM. Notably, \oursFramework with ChatGPT 0613 establishes a new state-of-the-art performance on \decision, surpassing the PaCE~\cite{li2023pace}. Interestingly, \oursFramework with GPT-4 1106 and LLaMa-2-Chat 70B exhibit lower performance, highlighting that the task of image sharing remains a challenging one for these models.

\input{tables/main_tables/few_shot_result}

\paragraph{Few-shot examples confuse \oursFramework in deciding the image-sharing behavior.} 
As shown in Table~\ref{tab:few_shot_main_result}, we evaluate the few-shot performance of \oursFramework with ChatGPT 1106 by varying the number of few-shot examples.  Interestingly, increasing the number of few-shot examples does not improve but rather decreases the \decision performance. Specifically, there is an inverse relationship between the number of few-shot examples and the performance of \decision. This finding contrasts with previous research on factual reasoning, such as in \cite{kojima2022large}, where few-shot examples are shown to improve the capabilities of LLMs. To thoroughly investigate this phenomenon, we measured the ratio of instances (\eg ``I'm sorry, I cannot assist with that request'') where \oursFramework refuses to make a decision (Refusal Ratio). As indicated in Table~\ref{tab:few_shot_main_result}, the refusal ratio significantly increases as performance decreases. This indicates that the presence of diverse decisions within few-shot examples confuses \oursFramework in making image-sharing decisions.

\input{tables/main_tables/decision_cot_result}

\paragraph{Chain-of-Thought is not effective in the image-sharing capability.}
Table~\ref{tab:cot_main_result} shows the zero-shot performance of \decision when we apply Chain-of-Thought (CoT) reasoning, specifically using the prompt ``Let's think step by step.'', as followed by~\citep{kojima2022large}. Generally, we observe that CoT does not enhance the performance of \decision. This suggests that inducing the model to think about the possibility of image-sharing behavior sequentially often leads to a conclusion in favor of sharing images. In fact, it is noted that the proportion of FP increases. These results confirm that the image-sharing behavior has significant subject property.

\input{tables/main_tables/retrieval_main_result}

\paragraph{\oursFramework enhances the dialogue-to-image retrieval task.}
Table~\ref{tab:retrieval_main_result} presents the zero-shot results of the dialogue-to-image retrieval task. Notably, \oursFramework, combined with various LLMs, significantly outperforms comparative models by a large margin of approximately 7.1\% (an absolute value). In addition, we observe that even large-scale pre-trained vision-and-language models (ViLT, PaCE, CLIP) underperform compared to human performance. This suggests that these models have limitations in comprehending dialogues and summarizing relevant image descriptions accurately. The image-sharing behavior also presents a considerable challenge to humans, underscoring its complexity.

Furthermore, we assess the image-sharing capability of recent large multi-modal models such as LLaVA v1.5, MiniGPT-4, and GPT4-V, which show remarkable performance on various visual-grounded language tasks (\eg VQA). Given that these models are not specifically designed for cross-modal retrieval tasks, we evaluate them by measuring the text similarity score between the generated image descriptions and dialogue history, using Sentence-BERT~\cite{reimers2019sentence}, as proposed in prior works. 
Among these models, GPT4-V achieves the best performance. However, their performance still falls short of that achieved by CLIP-large. This result indicates that, despite their versatility in visual-grounded language tasks, these models are not the optimal solution for tasks involving image-sharing behavior, likely due to their structural limitations.

\input{tables/main_tables/stage2_result}

\paragraph{\oursFramework has visual imagination ability better than humans.}
Table~\ref{tab:stage2_result} summarizes the zero-shot results from stage 2, focusing on \textsc{Descriptiveness}, \textsc{Completeness}, \textsc{Consistency}, and \textsc{Salient Information}. 
Contrasting with the \decision results, \oursFramework with GPT-4 1106 outperforms \oursFramework (w/ ChatGPT) in describing images, even surpassing human performance in \textsc{Descriptiveness} and \textsc{Completeness}. These findings support the notion that \oursFramework$_{\text{GPT-4 1106}}$ possesses notable visual imagination capabilities~\cite {lu2022imagination,zhu2023imagine,lu2023multimodal}. However, \oursFramework$_{\text{ChatGPT 0613}}$ demonstrates superior attention to \textsc{Salient Information} compared to \oursFramework$_{\text{GPT-4 1106}}$. This suggests that the task of image sharing remains a significant challenge for these models.

\begin{figure}[!t]
    \centering
    \includegraphics[width=\columnwidth]{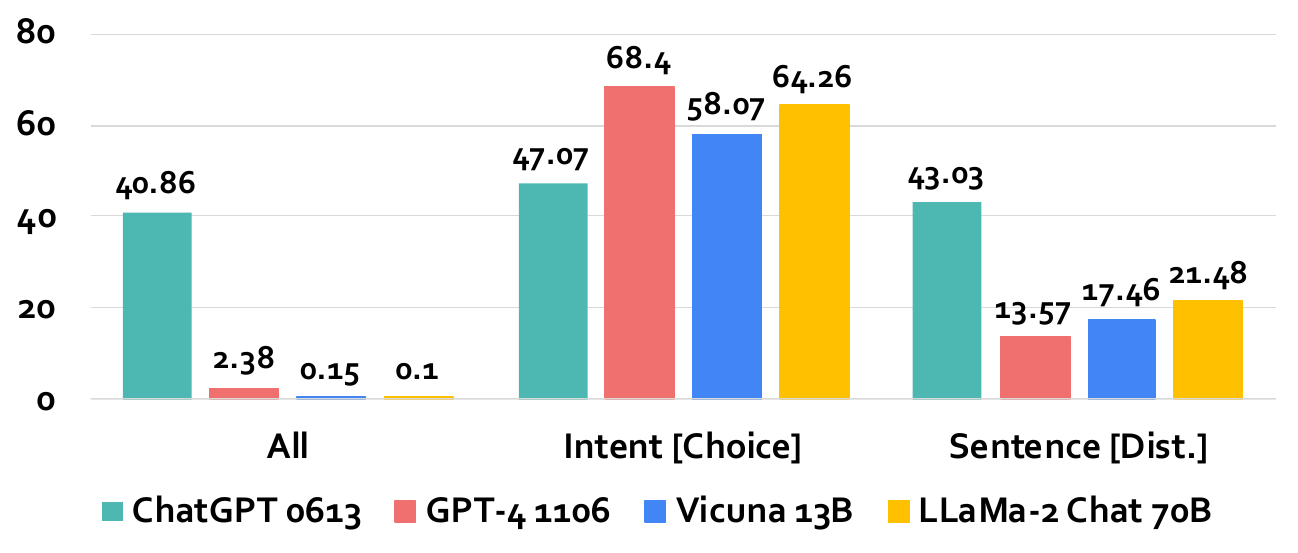}
    \caption{Results for \intent and \evidence are presented, with \textsc{All} signifying instances where \oursFramework correctly identifies \decision, \intent, and \evidence simultaneously.}
    \label{fig:intent}
    \vspace{-1em}
\end{figure}

\paragraph{Zero-shot Results of \intent.} 

Figure~\ref{fig:intent} shows the zero-shot results of \intent and \evidence on \photochatplus. Our findings reveal that \oursFramework with ChatGPT 0613 exhibits a lower performance in \intent than other models. However, regarding performance across \textsc{All} categories, \oursFramework with ChatGPT 0613 significantly outperforms the others. This observed decrease in \textsc{All} category performance among the other models indicates a selective impact on varying aspects of \decision, \intent, and \evidence.

\subsection{Applications} \label{sec:application}

We demonstrate the applicability and extensibility of our pipeline in two distinct applications: (1) Human-Bot Interaction, and (2) Dataset Augmentation.

\begin{figure}
    \centering
    \includegraphics[width=\columnwidth]{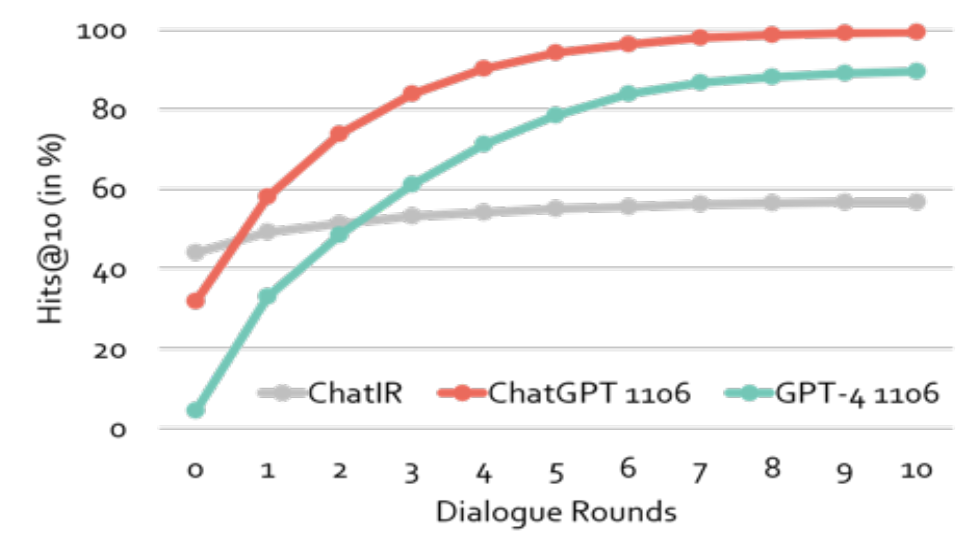}
    \caption{Zero-shot results of Hits@10 (\%) across multiple dialogue rounds are presented, comparing the ChatIR system~\cite{levy2023chatting} with \oursFramework. It should be noted that a dialogue comprising 0 rounds indicates that only the image description is provided to the model.}
    \label{main_fig:chatir_result}
    \vspace{-1em}
\end{figure}

\paragraph{Application 1: Human-Bot Interaction.} To validate the effectiveness of our pipeline in practical scenarios, we evaluate our pipeline on real human-bot interaction datasets based on VisDial~\cite{das2017visual} dataset, which is introduced in the prior work~\cite{levy2023chatting}. This specific scenario does not necessitate the decision-making process regarding image sharing. Therefore, \oursFramework is tailored to include only stages 2 and 3, showcasing its flexibility.
As shown in Figure~\ref{main_fig:chatir_result}, \oursFramework significantly outperforms the recent ChatIR system~\cite{levy2023chatting}, indicating its strong generalization performance. In addition, we find that the performance of \oursFramework is enhanced when provided with a more extensive dialogue context. These results indicate a strong and robust understanding of the dialogue context within \oursFramework, benefiting from the enormous ability of LLM.

\paragraph{Application 2: Dataset Augmentation.}

Previous studies~\citep{lee2021constructing,kim2022soda,chun2022eccv} have reported that machine-annotated or generated datasets significantly enhance generalization performance. Drawing inspiration from these works, we augment \photochat with \oursFramework equipped with ChatGPT 1106 by providing the full dialogue context to the LLM. The motivation for this is that given that the dialogue context in \photochat is already fixed, it is crucial to find image-sharing moments that wouldn’t disrupt the existing conversation flow after inserting relevant images at the image-sharing turns. Thus, we slightly modify the input prompt, as detailed in Appendix. Additionally, we prompt the LLM to generate a rationale for the act of image sharing, thereby gaining insight into the LLM's judgment. Once an image-sharing moment is identified, we align a relevant image to this moment using Stable Diffusion~\citep{rombach2022high}\footnote{\url{https://huggingface.co/stabilityai/stable-diffusion-2-1-base}}. This step is crucial as the LLM sometimes over-generates image descriptions, making it challenging for CLIP to align the relevant image using existing source image datasets like Conceptual Captions 3M~\cite{sharma2018conceptual}. We refer to this augmented dataset as \augphotochat.

\begin{figure}[t]
    \centering
    \includegraphics[width=\columnwidth]{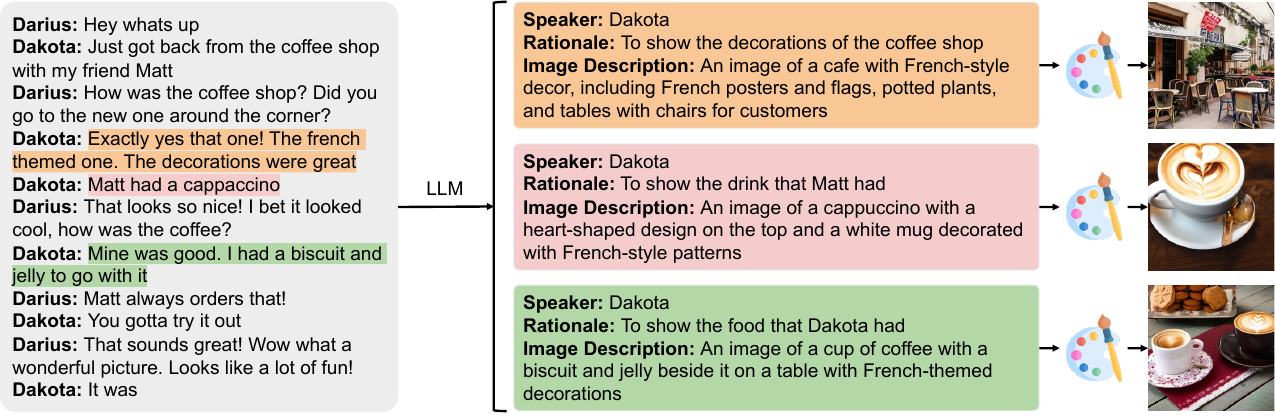}
    \caption{\textbf{An example of \augphotochat.} Followed by our pipeline, we construct the \augphotochat by generating appropriate images using Stable Diffusion (\drawingEmojiforcaption) by prompting predicted image descriptions from LLM (\ie ChatGPT 1106), which are highlighted in \textcolor{orangeforcaption}{orange}, \textcolor{pinkforcaption}{pink}, and \textcolor{greenforcaption}{green} boxes.} 
    \label{fig:case}
    \vspace{-1em}
\end{figure}

To assess if \augphotochat improves generalization performance on unseen multi-modal dialogue datasets, we implement simple text and image retrieval models (details in Appendix~\ref{supp_sec:finetune_expr}). We train these models on both PhotoChat and \augphotochat and evaluate them on PhotoChat and MMDialog~\citep{feng2022mmdialog}, a million-scale multi-modal dialogue dataset. Table~\ref{tab:aug_result} demonstrates that models trained on \augphotochat outperform those trained on the two multi-modal dialogue datasets. Notably, there is a significant performance boost on MMDialog with \augphotochat, highlighting the effectiveness of our pipeline in constructing \augphotochat. An example from \augphotochat is presented in Figure~\ref{fig:case}.

\input{tables/main_tables/aug_reuslt}

%% file: tables/main_tables/decision_main_result.tex
\begin{table}[t]

    \centering
    \begin{adjustbox}{width=0.8\columnwidth}
    
    \begin{tabular}{@{}lccc@{}}
    \toprule
    Model               & F1   & Precision & Recall \\ \midrule
    \multicolumn{4}{l}{\bluecell \textit{Fine-tuned Model}}       \\
    ALBERT-base         & 52.2 & 44.8      & 62.7   \\
    BERT-base           & 53.2 & 56.1      & 50.6   \\
    T5-base             & 58.1 & 58.2      & 57.9   \\
    T5-3B               & 58.9 & 54.1      & 64.6   \\
    ViLT                & 52.4 & 55.4      & 58.9   \\
    PaCE                & 63.8 & 63.3      & 68.0   \\
    \multicolumn{4}{l}{\bluecell \textit{Ours}}                       \\
    \oursFramework$_{\text{ChatGPT 0613}}$  &	\textbf{65.6} &	\textbf{66.7} &	64.7 \\
    \oursFramework$_{\text{ChatGPT 1106}}$        & 64.0 & 63.1      & 65.1   \\
    \oursFramework$_{\text{GPT-4 1106}}$          & 45.6 & 56.5      & \textbf{68.1}   \\
    \oursFramework$_{\text{Vicuna-13B}}$          & 65.5 & 64.3      & 67.1   \\
    \oursFramework$_{\text{LLaMa2-Chat-70B}}$     & 40.9 & 55.0      & 63.3   \\
    \oursFramework$_{\text{Falcon-Instruct-40B}}$ & 62.8 & 62.4      & 63.3   \\ \bottomrule
    \end{tabular}
    \end{adjustbox}

    \caption{Zero-shot results of \decision. The performance of fine-tuned models is reported from the previous work~\cite{li2023pace}. }
    
    \label{tab:decision_main_result}
    \vspace{-1em}
\end{table}

%% file: tables/main_tables/few_shot_result.tex
\begin{table}[t]

    \centering
    \begin{adjustbox}{width=\columnwidth}
    
\begin{tabular}{@{}lcccc@{}}
\toprule
Model                 & F1            & Precision     & Recall        & Refusal Ratio \\ \midrule
\oursFramework$_{\text{ChatGPT 1106}}$ (0 shot) & \textbf{64.0} & \textbf{63.1} & \textbf{65.1} & 0.0           \\
\oursFramework$_{\text{ChatGPT 1106}}$ (1 shot) & 60.8          & 60.5          & 61.0          & 0.1           \\
\oursFramework$_{\text{ChatGPT 1106}}$ (2 shot) & 37.7          & 53.8          & 59.4          & 4.2           \\
\oursFramework$_{\text{ChatGPT 1106}}$ (4 shot) & 27.2          & 53.5          & 55.5          & 10.6          \\
\oursFramework$_{\text{ChatGPT 1106}}$ (8 shot) & 26.5          & 53.7          & 54.8          & \textbf{20.4} \\ \bottomrule
    \end{tabular}
    \end{adjustbox}

    \caption{Few-shot results of \decision in \oursFramework, combined with ChatGPT 1106, by varying the number of few-shot examples. The Refusal Ratio (\%) indicates the ratio of generated answers where \oursFramework refuses to provide the decision.}
    
    \label{tab:few_shot_main_result}
    \vspace{-1em}
\end{table}

%% file: tables/main_tables/decision_cot_result.tex
\begin{table}[t]
    \centering
    \begin{adjustbox}{width=\columnwidth}
    \begin{tabular}{@{}lccccc@{}}
\toprule
Model              & F1   & Precision & Recall & FN  & FP   \\ \midrule
\oursFramework$_{\text{ChatGPT 0613}}$       & 65.6 & 66.7      & 64.7   & 629 & 499  \\
\oursFramework$_{\text{ChatGPT 0613}}$ + CoT & 65.1 & 65.1      & 65.2   & 606 & 617  \\ \cdashline{1-6}\noalign{\vskip 0.5ex}
\oursFramework$_{\text{ChatGPT 1106}}$       & 64.0 & 63.1      & 65.1   & 593 & 758  \\
\oursFramework$_{\text{ChatGPT 1106}}$ + CoT & 47.6 & 53.8      & 60.7   & 359 & 3648 \\ \bottomrule
    \end{tabular}
    \end{adjustbox}

    \caption{Zero-shot results of \decision in \oursFramework, combined with two different versions of ChatGPT when CoT is applied.}
    
    \label{tab:cot_main_result}
    \vspace{-1em}
\end{table}

%% file: tables/main_tables/retrieval_main_result.tex
\begin{table}[!t]
    \centering
    \begin{adjustbox}{width=0.8\columnwidth}
    \begin{tabular}{@{}lcccc@{}}
\toprule
Model                 & R@1                  & R@5                  & R@10                 & MRR                  \\ \midrule
Human & 19.6 & 37.5 & 44.7 & - \\
\multicolumn{5}{l}{\bluecell \textit{Fine-tuned Performance}}                                                                         \\
BM25                  & 6.6                  & 15.4                 & 23.0                 & -                    \\
DE                    & 9.0                  & 26.4                 & 35.7                 & -                    \\
VSE++                 & 10.2                 & 25.4                 & 34.2                 & -                    \\
SCAN                  & 10.4                 & 27.0                 & 37.1                 & -                    \\
VLMo                  & 13.8                 & 30.0                 & 39.4                 & -                    \\
ViLT                  & 11.5                 & 25.6                 & 33.8                 & -                    \\
PaCE                  & 15.2                 & 36.7                 & 49.6                 & -                    \\
DialCLIP & 19.5 & 44.0 & 55.8 & - \\
\multicolumn{5}{l}{\bluecell \textit{VLM, zero-shot}}                                                                                 \\
CLIP-base             & 13.7                 & 28.0                 & 35.2                 & 20.8                 \\
CLIP-large            & 14.1                 & 28.7                 & 35.3                 & 21.5                 \\
\multicolumn{5}{l}{\bluecell \textit{Large Multi-Modal Model}}                                                       \\
LLaVA v1.5 7B         & 11.1                 & 26.5                 & 33.3                 & 18.8                 \\
LLaVA v1.5 13B        & 12.1                 & 25.6                 & 32.3                 & 19.3                 \\
MiniGPT-4$_{\text{Vicuna 7B}}$  & 11.6                 & 26.5                 & 34.0                 & 19.1                 \\
MiniGPT-4$_{\text{Vicuna 13B}}$ & 11.7                 & 27.7                 & 35.5                 & 19.8                 \\
Qwen-VL-Chat 7B	& 12.1	& 27.4	& 36.1	& 20.2 \\
GPT4-V                & 13.8                 & 27.9                 & 35.9                 & 21.3                 \\
\multicolumn{5}{l}{\bluecell \textit{Ours}}                                                                                         \\
\oursFramework$_{\text{ChatGPT 0613}}$          & 26.6                 & 46.1                 & 54.2                 & 36.0                 \\
\oursFramework$_{\text{ChatGPT 1106}}$          & 26.3                 & 45.6                 & 54.3                 & 35.4                 \\
\oursFramework$_{\text{GPT-4 1106}}$           & \textbf{28.3}                 & \textbf{47.4}                 & \textbf{55.2}                 & \textbf{37.6}                 \\
\oursFramework$_{\text{Vicuna-13B}}$            & 25.8                 & 45.0                 & 53.1                 & 35.0                 \\
\oursFramework$_{\text{LLaMa2-Chat-70B}}$       & 24.5                 & 43.5                 & 52.6                 & 34.0                 \\ \bottomrule

    \end{tabular}
    \end{adjustbox}

    \caption{Zero-shot results of \oursFramework with various LLMs on the image retrieval task. The performance of fine-tuned models is reported from the previous work~\cite{li2023pace}.}
    
    \label{tab:retrieval_main_result}
    \vspace{-1em}
\end{table}

%% file: tables/main_tables/stage2_result.tex
\begin{table}[t]
\centering
\begin{adjustbox}{width=0.9\columnwidth}
\begin{tabular}{@{}lcccc@{}}
\toprule
Model        & Descriptiveness & Completeness & Consistency & Salient \\ \midrule
Human        & 0.1895          & 15.62        & -           & -        \\ 
\cdashline{1-5}\noalign{\vskip 0.5ex}
\oursFramework$_{\text{ChatGPT 0613}}$ & 0.1846          & 19.94        & \textbf{0.62}      & \textbf{0.1288}   \\
\oursFramework$_{\text{ChatGPT 1106}}$ & 0.1873          & 19.48        & \textbf{0.62}      & 0.0836   \\
\oursFramework$_{\text{GPT-4 1106}}$   & \textbf{0.1988}          & \textbf{20.80}         & 0.61      & 0.1013   \\ \bottomrule

\end{tabular}

\end{adjustbox}
\caption{Zero-shot results of stage 2.}
\label{tab:stage2_result}

\end{table}

%% file: tables/main_tables/aug_reuslt.tex
\begin{table}[t]
\centering
\begin{adjustbox}{width=\columnwidth}
\begin{tabular}{@{}lccccccc@{}}
\toprule
Eval   $\rightarrow$                 & \multicolumn{3}{c}{PhotoChat}                                                                          & \multicolumn{3}{c}{MMDialog}                                                              \\ \cmidrule{2-7}
Train      $\downarrow$              & \multicolumn{1}{c}{R@1} & \multicolumn{1}{c}{R@5} & \multicolumn{1}{c}{R@10} &  R@1                  & R@5                  & R@10                            \\ \midrule
 \multicolumn{7}{l}{\bluecell \textit{Image Retrieval}}       \\
PhotoChat                & 16.51$_{\pm 0.27}$          & 43.37$_{\pm 0.82}$          & 60.44$_{\pm 1.59}$          & 5.88$_{\pm 0.21}$          & 19.21$_{\pm 0.92}$          & 29.95$_{\pm 1.3}$            \\
\augphotochat              & \textbf{16.92$_{\pm 1.18}$} & \textbf{44.89$_{\pm 1.18}$} & \textbf{61.77$_{\pm 1.05}$} & \textbf{8.25$_{\pm 0.47}$} & \textbf{24.95$_{\pm 1.21}$} & \textbf{37.0$_{\pm 1.65}$}  \\ \midrule
\multicolumn{7}{l}{\bluecell \textit{Next Response Prediction}}           \\
PhotoChat                & 6.02$_{\pm 0.26}$           & 19.81$_{\pm 0.78}$          & 31.67$_{\pm 1.68}$          & 2.34$_{\pm 0.24}$          & 9.27$_{\pm 0.89}$           & 16.22$_{\pm 1.25}$        \\
\augphotochat              & \textbf{6.43$_{\pm 0.93}$}  & \textbf{23.06$_{\pm 1.52}$} & \textbf{34.24$_{\pm 1.15}$} & \textbf{2.67$_{\pm 0.07}$} & \textbf{9.86$_{\pm 0.26}$}  & \textbf{16.29$_{\pm 0.38}$}   \\ \bottomrule

\end{tabular}

\end{adjustbox}
\caption{We report the text and image retrieval performance across five runs on three multi-modal dialogue datasets: \photochat and MMDialog~\cite{feng2022mmdialog}.}
\label{tab:aug_result}

\end{table}

%% file: sections/06_related_work.tex
\section{Related Work} \label{sec:related_work}

\paragraph{Multi-Modal Dialogue Dataset.}

Existing studies predominantly fall into two categories, depending on whether the image in the dialogue is \textit{grounded} or \textit{sharing}. Image-grounded dialogue tasks are designed to answer questions~\citep{antol2015vqa,das2017visual,kottur2019clevr} or generate natural conversations~\citep{shuster2018image,meng2020openvidial,wang2021openvidial,zheng2021mmchat} about given images.
Nevertheless, it is common to share images pertinent to dialogue contexts in everyday conversations for the purpose of reinforcing social bonding, as well as enhancing engagement and interest.
Recent studies have proposed datasets that encapsulate this image-sharing behavior. This has been achieved by collecting a human-human dialogue dataset (PhotoChat) via crowdsourcing~\citep{zang2021photochat}, a large-scale dataset (MMDialog) from social media~\citep{feng2022mmdialog}, or constructing datasets automatically using vision-and-language models~\citep{lee2021constructing}.
In this work, our focus is exclusively on the PhotoChat dataset to gain a deeper understanding of the \textit{image-sharing} capabilities of LLMs. We do not include automatically constructed datasets or the MMDialog due to the considerable expense associated with conducting experiments using LLMs.

\vspace{-0.5em}

\paragraph{Prompting Large Language Models.}

Recent studies have witnessed the success of large language models, such as Instruct GPT-3~\citep{ouyang2022training}, ChatGPT~\citep{chatgpt}, GPT-4~\citep{openai2023gpt}, in a zero-/few-shot performance.
The use of these models, in conjunction with ``prompt engineering,'' has unlocked the abilities of LLMs, even emergent ones~\citep{wei2022emergent}, across various tasks. These tasks range from dialogues~\citep{lee2022does,kim2022soda,lee2022personachatgen}, complex reasoning tasks~\citep{wei2022chain,kojima2022large}, and theory-of-mind~\citep{sap2022neural,kosinski2023theory}, to image classification\citep{yang2022language,pratt2022does,menon2022visual,zhang2023prompt} and multi-modality~\citep{lu2023multimodal,han2023champagne}.

\paragraph{Large Multi-Modal Models.}

Recent studies have proposed large multi-modal models that demonstrate the surprising generalization performance on various visual-grounded language tasks, such as Flamingo~\cite{alayrac2022flamingo}, LLaVA~\cite{liu2023visual}, MiniGPT-4~\cite{zhu2023minigpt}, Qwen-VL~\cite{bai2023qwen}, CogVLM~\cite{wang2023cogvlm}, CoLLaVO~\cite{lee2024collavo}, MoAI~\cite{lee2024moai}, Meteor~\cite{lee2024meteor}, TroL~\cite{lee2024trol} and GPT4-V~\cite{gpt4v}.
In this work, we assess the image-sharing capability of these models in a zero-shot setting.

%% file: sections/07_conclusion.tex
\section{Conclusion} \label{sec:conclusion}

In this paper, we explore the \textit{image-sharing} capabilities of LLMs in a zero-shot prompting by introducing a three-stage framework: Decide, Describe, Retrieve (\oursFramework). 
Our extensive experiments demonstrate the effectiveness of our framework in enhancing zero-shot performance across both stages, with ChatGPT achieving state-of-the-art performance.
We also reveal that the \textit{image-sharing} ability is an emergent ability in the zero-shot prompting.
With extensive experiments, we reveal the effectiveness of our framework on two useful applications.
In future works, we will assess this capability in a few-shot setting using additional multi-modal dialogue datasets. We will construct a personalized multi-modal dialogue dataset using our framework.

\section*{Limitations} \label{supp_sec:limitation}

Here, we highlight some limitations of our work.
Firstly, our prompt template is rather lengthy, which complicates expansion into the few-shot setting. We anticipate that conducting few-shot prompting to utilize the image-sharing capability of LLMs would result in better performance compared to zero-shot prompting.
Secondly, LLMs tend to over-generate image descriptions even in the absence of specific demographic information such as age or appearance. For instance, in the description, ``An image of a woman with long, brown hair wearing a flowy white dress and brown boots,'' there is no reference to long hair in the given dialogue. Providing additional information (\eg persona) can enhance the relevance of image descriptions generated by LLMs.

\section*{Ethical Considerations} \label{supp_sec:broader_impact}

We report several potential issues with our proposed framework. 
First, generated image descriptions may propagate social bias because GPT-3 can still produce harmful content, including social bias and offensiveness\citep{baheti2021just,hartvigsen2022toxigen}.
Second, this issue has resulted in the inclusion of problematic descriptions in the constructed \augphotochat dataset, leading to socially-biased images generated using Stable Diffusion~\citep{rombach2022high}.
As a result, when vision-and-language models like CLIP~\citep{radford2021learning} and DALL-E~\citep{ramesh2021zero} are trained on this augmented dataset, they may exhibit social biases.
As reported in~\citep{wang2021gender}, even if we give the gender-neutral query to CLIP~\citep{radford2021learning} model, the CLIP model sometimes retrieves images causing gender-bias issues. We are concerned that this problematic issue may exist in the augmented dataset.
Therefore, the image retrieval model trained on this dataset may sometimes retrieve biased images. 
In addition, text-to-image generative models learn social biases from the augmented dataset, as reported in the prior work~\citep{cho2022dall}.
We should consider this problem important when building a multimodal search model.

%% file: sections/supplementary/00_human-evaluation.tex
\section{Details of Human Annotation Procedure} \label{supp:annotation_procedure}

\subsection{Preparing Human Annotation}
We first prepare 968 dialogues from the test set of \photochat to collect additional information crucial for understanding image-sharing behavior in a manner akin to human comprehension.
Then, we ask 12 human annotators to annotate various elements at the ground-truth moments of image-sharing in \photochat. These elements include \textit{intent}, \textit{triggering sentence}, \textit{salient information}, and \textit{image description}. 
Notably, we intentionally do not show the ground-truth images from \photochat during this annotation process. This process is adopted to explore the complexity and subjectivity inherent in image-sharing behavior.

\section{Human Evaluation Questionnaire}
This section lists the questions and notes used for human evaluations.

\subsection{Questions}
\begin{itemize}
    \item \textbf{What is the intent behind sharing the image?} 
    
        \textbf{Options:}
        Information Dissemination / Social Bonding / Human and Entertainment / Visual / Clarification / Topic Transition / Expression of Emotion or Opinion

        \textbf{Notes:} 
        You can select all of the following options that are possible. If the answer you have in mind is not listed, please write it in the space.
        If you write more than two sentences, then please divide them using the separator “;”.

    \item \textbf{Which sentence invokes the image-sharing behavior?} 

        \textbf{Notes:}
        You should highlight the most contributed sentence (i.e., only one sentence) using “Sentence”

    \item \textbf{What kind of image would be appropriate to share at the [Sharing Image] turn?} 
    
        \textbf{Notes:}
        You can write only one image description, starting with “An image of” or “A photo of”.

    \item \textbf{What words or phrases do you focus on to write the image description?} 
    
        \textbf{Notes:}
        You can highlight all words or phrases that you focus on using “Word/Phrase”
\end{itemize}

\begin{figure}
    \centering
    \includegraphics[width=\columnwidth]{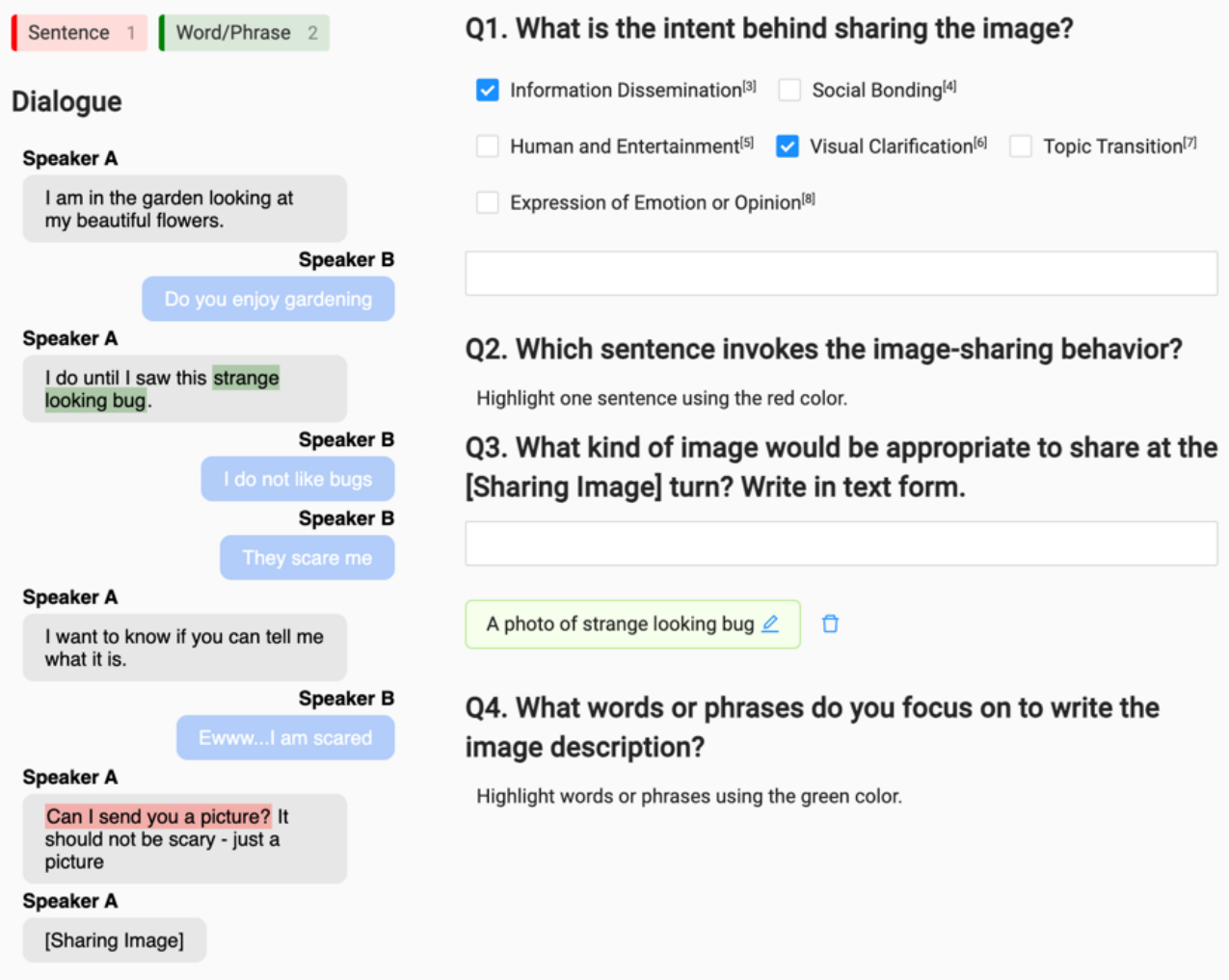}
    \caption{A screenshot of the human evaluation system for the human ratings.}
    \label{fig:label-studio}
\end{figure}

\section{Human Evaluation System}
We show a screenshot of the human evaluation system in Figure . We implement this system using Label Studio \cite{LabelStudio}.

\section{Details of Human Evaluation}
We recruited 12 individuals, unknown to us, who are either graduate or undergraduate students. Prior to participating in the experiment, they were provided with comprehensive instruction on the task, an overview of the multi-modal dialogue dataset, and a detailed explanation of the evaluation criteria. This preparatory phase lasted approximately one hour.

%% file: sections/supplementary/01_intrinsic.tex
\section{Discussions}

\paragraph{Domain Generalizability.} Towards Artificial General Intelligence (AGI), we need to show the proposed methods' generalization capability across different domains, such as medical imaging. We recognize the significance of this aspect and have designed our methodology to be both extensible and generalizable. Our approach's extensibility allows us to tailor \oursFramework according to specific domains. For instance, in a specialized domain like medical, we could incorporate a domain-specific LLM, such as Med-PaLM 2~\cite{singhal2023towards}, or even GPT-4, which has demonstrated superior performance in tasks requiring medical knowledge~\cite{nori2023capabilities}, including outperforming domain-specific models like Med-PaLM~\cite{singhal2023large} in the United States Medical Licensing Examination (USMLE). For the image retrieval aspect, a specialized model such as MedCLIP~\cite{wang2022medclip} could be employed to enhance retrieval accuracy in the medical context.

\paragraph{Towards Better Image-Sharing Ability.}

As shown in our experiment, the likelihood of performance improvement is high as the model's size increases or when it is trained with alignment to human preference. This suggests that the image-sharing ability is subjective and resembles human-like tasks. Therefore, receiving real-time feedback through interactive mode (a form of human-AI collaboration) and further training the model using the RLHF method could lead to better performance, aligning the model's actions favorably with image-sharing ability.

Furthermore, understanding conversational context is essential, and imbuing the model with the ability of perspective-taking, understanding situations from the user's point of view, could lead to performance enhancement. For instance, when a user is feeling down due to poor test results, the model could not only provide empathy through text but also share a picture of a dog based on the user's fondness for dogs and the current context of struggling with test scores, thereby offering multi-faceted empathy.

In addition, unlike image-grounded dialogue, image-sharing scenarios might lack explicit information from previous conversations. For instance, understanding what ``it'' refers to in ``I love it'' requires considering the preceding conversational context. Thus, it's important to consider coreference resolution. Moreover, while sharing images, incorporating information about significant utterances from previous dialogues or using keywords and keyphrases could likely improve performance.

As depicted in Figure~\ref{fig:case}'s orange-generated results, the language model might sometimes over-generate due to excessive creativity. For instance, if the conversation only contains information about a coffee shop without mentioning ``French-style,'' the model might still produce the word ``French.'' Such cases could pose challenges in practical applications where inappropriate images could be retrieved.

In practical applications, it's beneficial to consider the user's satisfaction and share images that account for their personal information. For example, if a user mentions, ``I work in a hotdog stand,'' and their friend, who also works there, has a picture related to selling hotdogs in their phone album, it would be more suitable to share an image depicting the user themselves selling hotdogs rather than an image with the friend. Of course, obtaining explicit consent for sharing personal information is crucial.

Additionally, beyond improving the image-sharing ability, at the application level, using videos could enhance user engagement. Exploring this avenue could be a promising direction for future research.

\paragraph{Intrinsic Properties of LLMs.}

We believe that the intrinsic properties of LLM, which have been experimentally proven in various studies, have influenced image-sharing ability.

\begin{itemize}
    \item \textbf{Understanding the dialogue context:} It's essential to grasp the conversation topic holistically, emotional shifts between users, and general knowledge. Recent research results have shown that language models possess these abilities.
    \item \textbf{Understanding the interlocutor's mental state:} It is important to comprehend the interlocutor is situation to determine whether sharing an image is appropriate. For instance, if the interlocutor is upset, it might be better to respond empathetically rather than share an image. This ability is highly related to the Theory-of-Mind (ToM). Recently, LLMs have achieved competitive performance in Theory-of-Mind (ToM) tasks, which may influence image-sharing ability.
    \item \textbf{Understanding the intent:} From the model's perspective, sharing an image can be seen as intent. Many language models have demonstrated good performance in task-oriented dialogue tasks.
    \item \textbf{Visual imagination ability:} To share an appropriate image, one must imagine which image is best. This capability has been empirically proven in various recent studies. We investigated the C4 dataset, a representative pretraining dataset for LLMs, to analyze why this capability is manifested. The data discovered in C4 consists of pairs of images and their corresponding captions. These captions contain words/phrases related to visual imagination ability, such as ``depict'' and ``photo of.'' Moreover, on blogs, images often appear consecutively along with stories. Due to these elements, the LLM learned an inherent visual, and imaginative capability during its pretraining phase.
\end{itemize}

%% file: sections/supplementary/02_implementation_details.tex
\section{Implementation Details of LLMs} \label{supp_sec:implementation}

To evaluate the \textit{image-sharing} capabilities of LLMs, we call ChatGPT~\citep{chatgpt} and GPT-4~\citep{openai2023gpt} by calling OpenAI API.
All experiments are conducted on two A100 (40GB) GPU.
For each stage, the generation configuration is as follows:

\begin{itemize}
    \item For Stage 1, we set maximum tokens to 1024, temperature to 0.0, frequency penalty to 0.0, presence penalty to 0.0, top\_p to 1.0, and stop tokens to \texttt{\textbackslash n\textbackslash n}.
    \item For Stage 2, we set maximum tokens to 1024, temperature to 0.9, frequency penalty to 0.0, presence penalty to 0.4, top\_p to 0.95, and stop tokens to a default setting.
\end{itemize}

\section{Details of Experimental Settings} \label{supp_sec:finetune_expr}

To explore how our dataset affects both text and image retrieval tasks, we implement two simple and standard baseline retrieval models for text-to-image and image-to-text settings. 

\subsection{Task Definition}

Follwing~\citep{lee2021constructing,zang2021photochat}, we explain the formulation of two main tasks - next response prediction and image retrieval. 
Let us assume that we have a multi-modal dialogue $\mathcal{D} = \{(u_j, i_j, c_j)\}_1^N$ where $N$ denotes the number of dialogue turns, and $j=t$ is the turn that an image sharing behavior occurs. Then, each task is formulated as follows.

\paragraph{Next response prediction} is to predict the next utterance at turn $t+1$ given the dialogue history ($\{u_j\}_1^t$) and image $i_t$.

\paragraph{Image retrieval} is to retrieve relevant image at turn $t$ given the dialogue history ($\{u_j\}_1^{t-1}$).

Following~\citep{shuster2018image,lee2021constructing}, we set the number of retrieval candidates to 100 and use Recall@\{1,5,10\} and mean reciprocal rank (MRR) for the evaluation metrics.

\begin{figure}
    \centering
    \includegraphics[width=\linewidth]{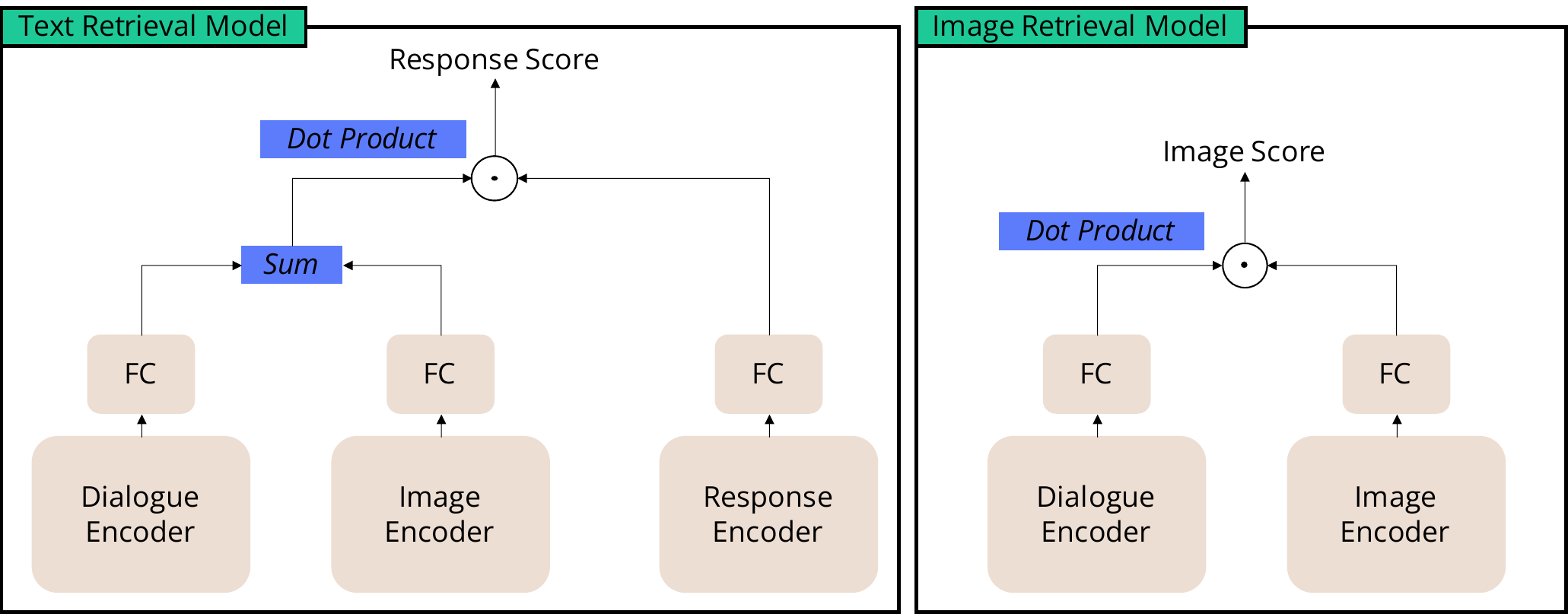}
    \caption{Architectures of two baseline models: Text retrieval and Image retrieval.}
    \label{supp_fig:basline}
\end{figure}

\subsection{Baseline Models} \label{supp_sec:baselines}

As illustrated in Figure~\ref{supp_fig:basline}, we present the architecture of baseline models: the text retrieval and image retrieval models. We provide a detailed description of baseline models below.

\paragraph{Text Retrieval Model.}
The text retrieval model consists of three primary components: the dialogue encoder, the response encoder, and the image encoder. The dialogue encoder processes the entire dialogue history into a fixed-size representation using the BERT model~\citep{devlin2018bert}. The dialogue history includes up to three turns preceding the current turn, with each turn separated by the \texttt{[SEP]} token. The response encoder also converts responses into fixed-size representations, utilizing a different version of the BERT model than the dialogue encoder. After BERT processes the text, mean pooling is applied to the text representations for both encoders. These pooled representations then pass through a linear projection layer followed by the ReLU activation function~\citep{nair2010rectified}. The image encoder uses the CLIP-base model~\citep{radford2021learning} to extract feature vectors from images. The dialogue and image feature vectors are then combined using element-wise addition. The loss is computed by taking the dot product between the response feature vector and the resulting summed vector.

\paragraph{Image Retrieval Model.}
The image retrieval model is composed of two main components: the dialogue encoder and the image encoder. The dialogue encoder uses the BERT-base model to convert the dialogue into a representation, followed by mean pooling of the text representations. For the image representation, the CLIP-base model is utilized. After encoding, the image and dialogue vectors are passed through their respective linear projection layers, each followed by a ReLU activation function. The loss is determined by calculating the dot product between the image feature vector and the dialogue vector.

\subsection{Implementation Details}

We implement baseline models based on PyTorch Lightning. All experiments are conducted on two A100 GPUs (40GB). To accelerate the training time, we apply distributed training to baselines. We follow the hyperparameter settings similar to the previous works~\citep{lee2021constructing,zang2021photochat}, which are described as follows:

\paragraph{Text retrieval.} In our experiment, we set the batch size to 256, the learning rate to 5e-5, and the gradient clipping value to 2.0. We use the AdamW optimizer with a cosine learning rate scheduler. We set the warm-up ratio as 0.1\% and weight decay as 0.2.

\paragraph{Image retrieval.} We set the batch size to 256. We also use the AdamW optimizer with an initial learning rate of 2e-5 and decaying 0.1\%. 

\paragraph{Training.} Since our dataset contains several images per utterance, we randomly choose one image in each batch. We do not update the parameter of the image encoder. 

%% file: sections/supplementary/03_rationale_dist.tex
\section{Rationale Distribution} \label{supp_sec:rationale_dist}

We present the rationale distribution as shown in Table~\ref{supp_tab:rationale_dist}.

\input{tables/supp_tables/rationale_distribution}

%% file: tables/supp_tables/rationale_distribution.tex
\begin{table*}[ht]
\centering
\begin{adjustbox}{width=\textwidth}
\begin{tabular}{llrl}
\toprule
Verb & Object & Count & Example \\
\midrule
provide & information &    612 &                          To provide more information about the moth she saw. \\
         & context &    445 &                                     To provide context for the conversation. \\
         & representation &    397 &  To provide a visual representation of the beverage person is talking about. \\
         & evidence &    215 &               To provide visual evidence of the fun time they had together.  \\
\midrule
show & interest &    174 &                                    To show his interest in seeing the photo. \\
         & image &    173 &                   To show the image of the letters he formed with the dough. \\
         & person &    149 &                             To show person that he is okay with the weather. \\
         & audience &    111 &               To show the audience the fun person is having on his vacation. \\
\midrule
share & image &    145 &                                    To share the image of the birthday party. \\
         & photo &     13 &                                              To share the photo with person. \\
\midrule
express & interest &     30 &                                        To express interest in person's story \\
         & reaction &     27 &                                        To express her reaction to the image. \\
         & excitement &     17 &                                    To express excitement about the workshop. \\
         & appreciation &     12 &                                    To express her appreciation for the cake. \\
\midrule
invite & person &     38 &                            To invite person to see the picture of the table. \\
\midrule
ask & person &     25 &                       To ask person to share an image of his recent cooking. \\
         & question &      6 &                                 To ask a follow-up question about the image. \\
\midrule
encourage & person &     23 &                     To encourage person to share his most memorable dinner. \\
\midrule
introduce & image &     12 &                                                      To introduce the image. \\
         & topic &      8 &                                  To introduce the topic of the conversation. \\
\midrule
gauge & interest &     18 &                               To gauge person's interest in the baked goods. \\
\midrule
give & opportunity &     13 &                     To give person the opportunity to see a photo of Hannah. \\
\midrule
engage & person &      9 &    To engage person in the conversation and to show her the photo Zora sent. \\
\midrule
emphasize & importance &      8 &                      To emphasize the importance of spending time with kids. \\
\midrule
indicate & interest &      7 &                           To indicate person's interest in seeing the photo. \\
\bottomrule
\end{tabular}
\end{adjustbox}
\vspace{0.7pt}
\caption{\textbf{Rationale Distribution.} The top 20 most common root verbs and their up to 4 direct noun objects in the generated rationale. Only pairs with a count of 5 or more are included.}
\label{supp_tab:rationale_dist}
\end{table*}

%% file: sections/supplementary/04_template.tex
\section{Prompt Templates} \label{supp:prompt_template}

Here, we present all prompt templates used in our work, such as restriction-based prompt templates for each stage, and several prompt templates for the ablation studies.

\subsection{Prompt Templates}

We present a prompt template for dataset augmentation in our proposed framework, as shown in Figure~\ref{supp_fig:dataset_augmentation_prompt_template}.

\input{figures/supplementary_figures/prompt_template}

\input{figures/supplementary_figures/abl_prompt_template}

%% file: figures/supplementary_figures/prompt_template.tex
\setlength{\columnsep}{0.2cm}
\begin{figure*}[ht]

\begin{tcolorbox}[
    colback=qualcolor!5!white,
    colframe=qualcolor!75!black,]
\begin{small}
\textbf{Prompt Template for Stage 1:}\\
The following is a dialogue between \texttt{[speaker1]} and \texttt{[speaker2]}. You should share an image to make the following dialogue more interesting and engaging. The dialogue is provided line-by-line. In the given dialogue, select all utterances that are appropriate for sharing the image in the next turn, and write the speaker who will share the image after the selected utterance. You should also provide a rationale for your decision. \\
\\
Dialogue:\\
\texttt{[dialogue]} \\
\\
Restrictions: \\
(1) your answer should be in the format of "<UTTERANCE> | <SPEAKER> | <RATIONALE>". \\
(2) you MUST select the utterance in the given dialogue, NOT generate a new utterance. \\
(3) the rationale should be written starting with "To". \\
\\
Answer:\\
1.
\end{small}
\Sepline

\begin{small}
\textbf{Prompt Template for Stage 2:}\\
The following is a dialogue between \texttt{[speaker1]} and \texttt{[speaker2]}. The dialogue is provided line-by-line. \texttt{[speaker1]} shares an image in a given dialogue to make the following dialogue more interesting and engaging, marked in [Sharing Image]. Depict the most appropriate image to be shared in the next turn, in detail.\\
\\
Dialogue:\\
\texttt{[dialogue]}\\
\\
Restrictions:\\
(1) your answer should be written starting with "An image of" and in one sentence.\\
(2) you do NOT include the speaker's name (i.e., \texttt{[speaker1]}, \texttt{[speaker2]}) in the image description.\\
(3) you should share a realistic image, NOT memes.\\
\\
Image Description:
\end{small}

\end{tcolorbox}

\caption{\textbf{Prompt Templates for Dataset Augmentation.} A prompt template for stage 1 (\textbf{top}). A prompt template for stage 2 (\textbf{bottom}).}
\label{supp_fig:dataset_augmentation_prompt_template}
\end{figure*}

%% file: figures/supplementary_figures/abl_prompt_template.tex
\setlength{\columnsep}{0.2cm}
\begin{figure*}[ht]
\begin{tcolorbox}[
    colback=qualcolor!5!white,
    colframe=qualcolor!75!black,]
\begin{small}
\textbf{Prompt Template for Stage 1:}\\
The following is a dialogue between \texttt{[speaker1]} and \texttt{[speaker2]}. The dialogue is provided line-by-line. You will be provided a list of the intent of sharing an image. In the given dialogue, you should predict whether it is appropriate for \texttt{[share\_speaker]} to share an image in the next turn, the intent of the image-sharing behavior, and one sentence that invokes the image-sharing behavior. \\
\\
List of Intent of Image-Sharing Behavior:\\
- Information Dissemination: This involves sharing images to communicate important information, such as news, economic updates, or educational material (infographic image), aiming to inform or educate\\
- Social Bonding: This involves sharing images to strengthen social connections, including personal photos and memories\\
- Humor and Entertainment: This involves sharing light-hearted images, such as funny pictures or memes, to entertain and bring joy\\
- Visual Clarification: This involves sharing images, such as diagrams, item-specific photos, or location images, to clarify complex concepts or situations\\
- Topic Transition: This involves sharing images to shift the conversation topic or mood\\
- Expression of Emotion or Opinion: This involves sharing images, such as emotive photos or art, to express emotions or opinions more effectively than text, succinctly conveying feelings or perspectives\\
\\
Dialogue:\\
\texttt{[dialogue]}\\
\\
Question: Is it appropriate for \texttt{[share\_speaker]} to share an image in the next turn? If "Yes", choose all possible intents of sharing the image and provide only one sentence that invokes the image-sharing behavior.\\
Options:\\
(a) Information Dissemination\\
(b) Social Bonding\\
(c) Humor and Entertainment\\
(d) Visual Clarification\\
(e) Topic Transition\\
(f) Expression of Emotion or Opinion\\
Restrictions: \\
(1) You should provide your answer in a Python dictionary object with three keys, "Prediction", "Intent", and "Sentence".\\
(2) You should provide a binary answer (i.e., "yes" or "no") for the value of the "Prediction" key.\\
(3) You should choose all possible intents for the value of "Intent" key.\\
(4) You should provide the most contributed sentence (i.e., only one sentence) that invokes the image-sharing behavior for the value of "Sentence" key.\\
Answer:
\end{small}

\end{tcolorbox}

\caption{\textbf{Prompt Templates for Stage 1.} A prompt template used in the ablation study in Stage 1.}
\label{supp_fig:prompt_template_stage1}
\end{figure*}

\setlength{\columnsep}{0.2cm}
\begin{figure*}[ht]
\begin{tcolorbox}[
    colback=qualcolor!5!white,
    colframe=qualcolor!75!black,]
\begin{small}
\textbf{Prompt Template for Stage 2:}\\
The following is a dialogue between \texttt{[speaker1]} and \texttt{[speaker2]}. The dialogue is provided line-by-line. \texttt{[share\_speaker]} shares an image in a given dialogue to make the following dialogue more interesting and engaging, as marked in [Sharing Image]. \\
\\
Dialogue:\\
\texttt{[dialogue]}\\
\\
Question: What is the most appropriate image description to share in the [Sharing Image] turn?\\
Restrictions:\\
(1) You should provide your answer in a Python dictionary object with "Image Description" key.\\
Answer:
\end{small}

\end{tcolorbox}

\caption{\textbf{Prompt Templates for Stage 2.} A prompt template used in the in Stage 2.}
\label{supp_fig:prompt_template_for_stage2}
\end{figure*}

\begin{figure*}
\begin{tcolorbox}[
    colback=qualcolor!5!white,
    colframe=qualcolor!75!black,]

\begin{small}
\textbf{Prompt Template for Completeness:}\\
You will be provided a list of object categories and an image description. Your job is to detect the object in the image description and categorize the detected object into one of the categories in the list. \\
\\
You must provide your answer in a Python dictionary object that has the category as the key and the corresponding object in the image description as the value.\\
\\
Object Category List = [
    "Woman", "Man", "Girl", "Boy", "Human body", "Face", "Bagel", "Baked goods", "Beer", "Bread", "Burrito", 
    "Cake", "Candy", "Cheese", "Cocktail", "Coffee", "Cookie", "Croissant", "Dessert", "Doughnut", "Drink", 
    "Fast food", "French fries", "Hamburger", "Hot dog", "Ice cream", "Juice", "Milk", "Pancake", "Pasta", 
    "Pizza", "Popcorn", "Salad", "Sandwich", "Seafood", "Snack", "Animal", "Alarm clock", "Backpack", "Blender", 
    "Banjo", "Bed", "Belt", "Computer keyboard", "Computer mouse", "Curtain", "Guitar", "Hair dryer", "Hair spray", 
    "Harmonica", "Humidifier", "Jacket", "Jeans", "Dress", "Earrings", "Necklace", "Fashion accessory", "Bicycle", 
    "Calculator", "Camera", "Food processor", "Jug", "Mixing bowl", "Nightstand", "Oboe", "Oven", "Paper cutter", 
    "Pencil case", "Perfume", "Pillow", "Personal care", "Pizza cutter", "Pressure cooker", "Printer", "Refridgerator", 
    "High heels", "Skateboard", "Slow cooker", "Teddy bear", "Teapot", "Vase", "Wall clock", "Taco", "Tart", "Tea", 
    "Waffle", "Wine", "Guacamole"
]\\
\\
Image Description: \texttt{[description]}\\
\\
Answer:
\end{small}

\end{tcolorbox}
\caption{\textbf{Prompt Templates for \textsc{Completeness}.} A prompt template for evaluating \textsc{Completeness}.}
\label{supp_fig:prompt_template_for_completeness}
\end{figure*}

%% file: sections/supplementary/05_case_studies.tex
\section{More Examples of \augphotochat} \label{supp_sec:more_example}

We provide more examples of \augphotochat.

\begin{figure*}[h]
    \centering
    \includegraphics[width=\linewidth]{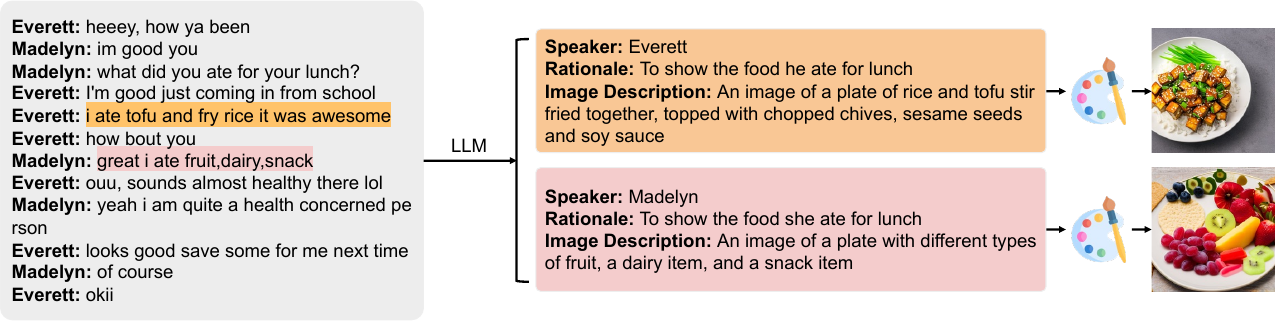}
    \includegraphics[width=\linewidth]{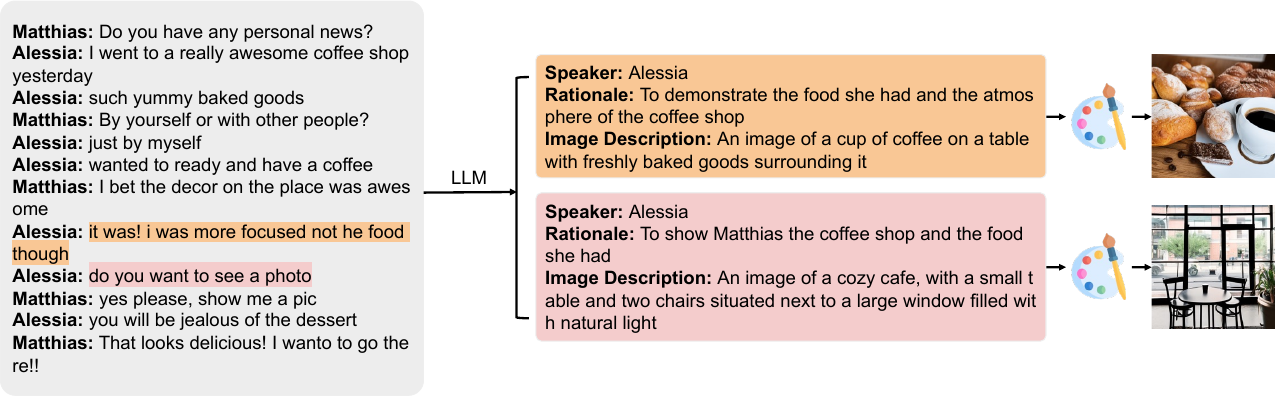}
    \includegraphics[width=\linewidth]{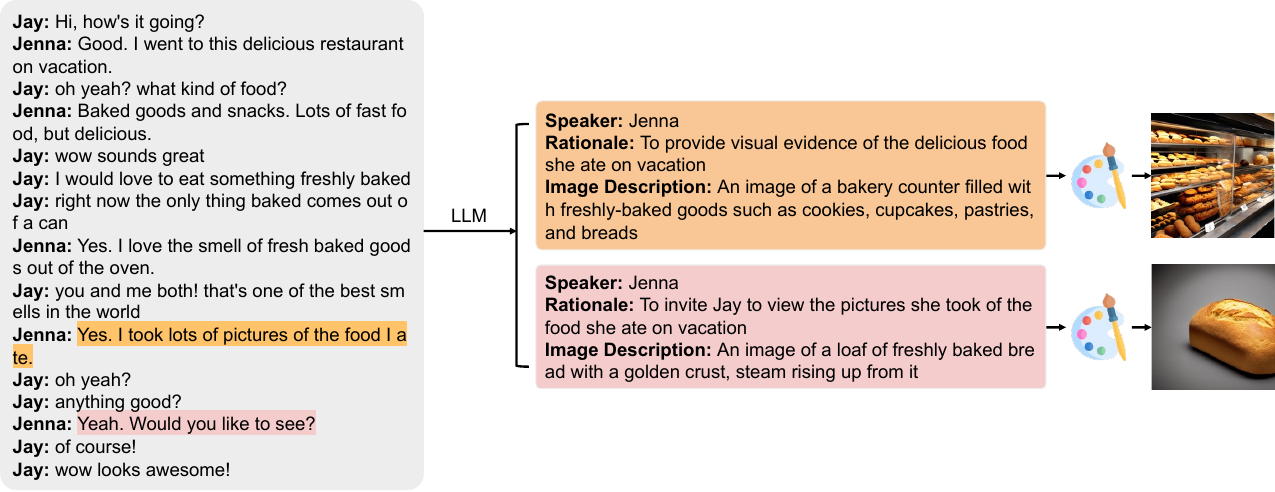}
    \includegraphics[width=\linewidth]{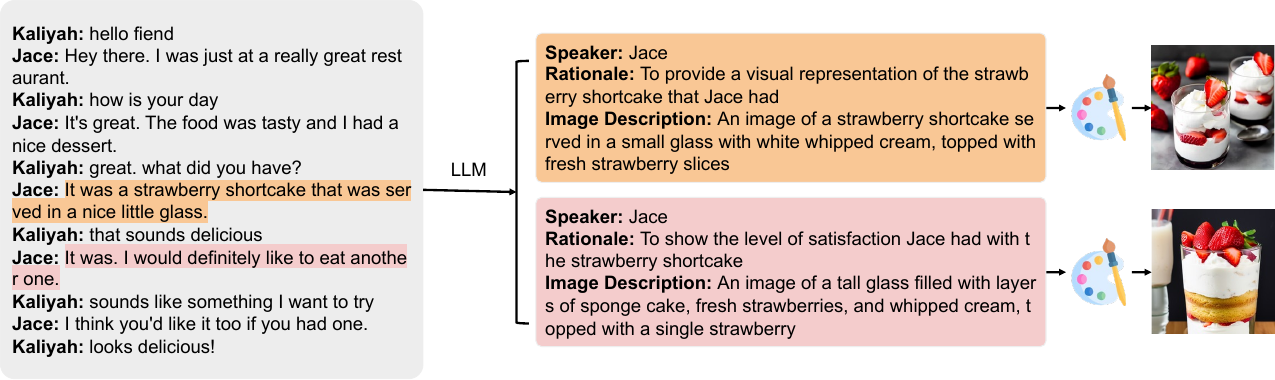}
    \includegraphics[width=\linewidth]{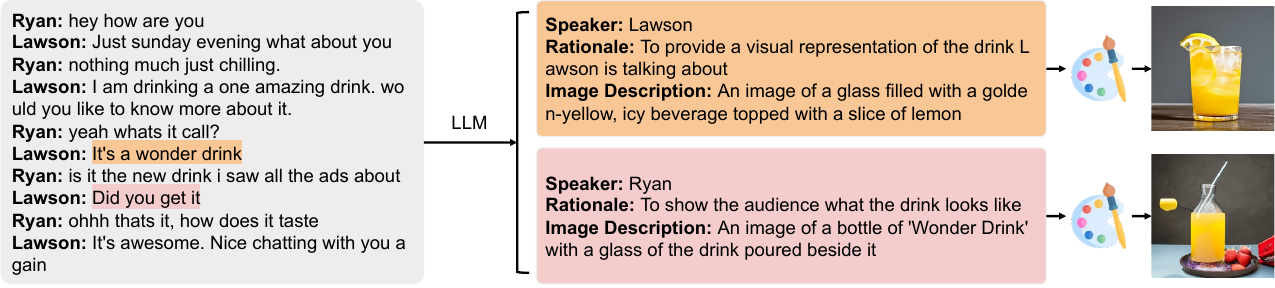}
    \caption{\textbf{More Examples of \augphotochat.} We present more generated examples of \augphotochat dataset using our proposed framework with LLM (\ie \texttt{ChatGPT 1106}) and Stable Diffusion(\drawingEmojiforcaption).}
    \label{supp_fig:more_example}
\end{figure*}